\documentclass{article}

\usepackage{arxiv}

\usepackage[utf8]{inputenc} 
\usepackage[T1]{fontenc}    
\usepackage{hyperref}       
\usepackage{url}            
\usepackage{booktabs}       
\usepackage{amsfonts}       
\usepackage{nicefrac}       
\usepackage{microtype}      
\usepackage{lipsum}         
\usepackage{caption} 
\usepackage{graphicx}
\usepackage{natbib}
\usepackage{doi}
\usepackage{physics}
\usepackage{lscape}
\usepackage{afterpage}
\usepackage{mdframed}
\usepackage{multirow}
\usepackage{xltabular}
\usepackage{booktabs}
\usepackage{ltablex}
\usepackage{enumitem}
\usepackage{supertabular}
\usepackage{subcaption}
\usepackage{mathtools}
\usepackage{amsmath}
\newtheorem{definition}{Definition}%
\DeclareMathOperator*{\argmax}{argmax}
\newtheorem{Example}{Example}[section]

\title{Directed Graph-alignment Approach for Identification of
Gaps in Short Answers}

\date{}

\newif\ifuniqueAffiliation
\uniqueAffiliationtrue

\ifuniqueAffiliation 
\author{{\hspace{1mm}Archana Sahu}\\
	Centre for Educational Technology\\
	Indian Institute of Technology Kharagpur\\
	West Bengal, India \\
	\texttt{sahuarchana7@gmail.com} \\
	\And
	{\hspace{1mm}Plaban Kumar Bhowmick} \\
	G.S. Sanyal School of Telecommunications\\
	Indian Institute of Technology Kharagpur\\
	West Bengal, India \\
	\texttt{plaban@cet.iitkgp.ac.in} \\
}

\else
\usepackage{authblk}

\author[1]{%
	{\hspace{1mm}Archana Sahu\thanks{\texttt{sahuarchana7@gmail.com}}}%
}
\author[1,2]{%
	{\hspace{1mm}Plaban Kumar Bhowmick\thanks{\texttt{plaban@cet.iitkgp.ac.in}}}%
}
\affil[1]{Centre for Educational Technology\\
	Indian Institute of Technology Kharagpur\\
	West Bengal, India \\}
\affil[2]{G.S. Sanyal School of Telecommunications\\
	Indian Institute of Technology Kharagpur\\
	West Bengal, India \\}
\fi

\renewcommand{\shorttitle}{Directed Graph-Alignment Approach for Identification of Gaps in Short Answers}

\hypersetup{
pdftitle={Directed Graph-Alignment Approach for Identification of Gaps in Short Answers},
pdfauthor={Archana Sahu, Plaban Kumar Bhowmick},
pdfkeywords={Automatic answer grading, NLP for education, Directed graph alignment, Similarity flooding},
}

\begin{document}
\maketitle

\begin{abstract}	
In this paper, we have presented a method for identifying missing items known as gaps in the student answers by comparing them against the corresponding model answer/reference answers, automatically. The gaps can be identified at word, phrase or sentence level. The identified gaps are useful
in providing feedback to the students for formative assessment.
The problem of gap identification has been modelled as an
alignment of a pair of directed graphs representing a student
answer and the corresponding model answer for a given question.
To validate the proposed approach, the gap annotated student
answers considering answers from three widely known datasets
in the short answer grading domain, namely, University of North
Texas (UNT) \citep{mohler2009text}, SciEntsBank \citep{dzikovska2012towards}, and Beetle \citep{dzikovska2011beetle} have been
developed and this gap annotated student answers' dataset is available at: https://github.com/sahuarchana7/gaps-answers-dataset. Evaluation metrics used in the traditional machine
learning tasks have been adopted to evaluate the task of gap
identification. Though performance of the proposed approach
varies across the datasets and the types of the answers, overall
the performance is observed to be promising.
\end{abstract}


\section{Introduction}
Assessment is an integral component of the teaching-learning process. The assessment of the student answers
is performed in either of the two forms: summative and
formative. In general, summative assessment is administered
to measure the achievement of the learners at the end of
the instruction delivery using a predefined scoring criteria.
Formative assessment, on the other hand, is administered with
the objective of providing qualitative feedback on the student
answers. Such feedback motivates the students to compare
their answers with the corresponding correct answers and to
identify any learning gaps.

Owing to the growth in the number of students in the
classrooms and the rapid adoption of the online learning
platforms like the Massive Open Online Courses (MOOCs),
the automatic grading of answers has become a very important
technology requirement in the education domain. The technology for automatic evaluation of the objective questions,
namely, Multiple Choice Questions (MCQs) or a related
family, is trivial. However, the MCQs often fail to assess
the conceptual understanding of the students. On the other
hand, the questions that elicit free-text answers from the
students provide them the freedom of expression. However,
the automatic assessment of the free-text answers is not that trivial compared to the assessment of the MCQs. The difficulty
in developing an automated assessment system for the free-text answers can be attributed to the linguistic variation of
the core concepts in the student answers. To address the
challenges, several automated answer grading systems \citep{DBLP:conf/semeval/HeilmanM13}, \citep{archplaban2019ensemble},
specifically for the short answers, have been proposed in the
last few decades \citep{Burrows2015}. Though over the time, the automated
short answer grading systems became reliable, thanks to the
benchmark datasets and the growing interest of the Natural
Language Processing community, these systems are useful for
the summative assessment setting only. However, these systems have limited application in formative assessment which
requires the gaps in the students answers to be identified. A
review of the automated answer evaluation literature revealed
that the research in the automated student answer analysis with
the objective of formative assessment is rather sparse.

In the present work, we aim at developing an unsupervised
model for the automatic identification of gaps (in word, phrase, or sentence level) in short answers provided by the students with a broader objective of providing formative feedback. The proposed system identifies the missing concepts (gaps) in the student answers by comparing them against one or more pre-specified model answers. 
The problem of identification of gaps in student answer as a part of providing formative assessment for the student answer \citep{arch2016identifygaps}, addressed within the scope of the present work, can be defined formally as follows: 
\begin{definition}{\textit{Identification of Gaps for Formative Assessment}} \\
Input: A pair of short answers $(M,~S)$ to a question $Q$ where: $M$ Model answer, 
$S$: Student answer to be analyzed \\
Output: A set $G_Q$ = $\{g_{1}, g_{2}, ..., g_{n}\}$
where $g_i$ denotes a gap in the student answer $S$ against the question $Q$.
\end{definition}

In an intuitive sense, it is quite possible that a student answer may capture some parts of the model answer, if not all. Considering a student answer and the corresponding model answer in parallel, fragments in the model answer that are present in the student answer can be aligned. The fragments in the model answer that have not been aligned with any component of the student answer can be considered as gaps. The approach, though very simple to implement, may not perform well as it does not consider the
semantic structure of the answers. This idea of word-word or chunk-based alignment approach has been implemented in \citep{arch2016identifygaps}. 

The word-word alignment system operates as a sequence of alignment modules, where each module aligns different types of word-pairs such as identical word sequences, named entities, content words, stop words. Decisions regarding alignment are taken by each module, after taking into account contextual evidence. The mapping between dependency types obtained from dependency parses for a pair of answers provides contextual evidence for alignment. The semantic similarity between neighboring word-pairs of the word-pair considered for alignment also contributes towards contextual evidence. It was hypothesized that analysis based on the semantic structure of the answers may lead to a superior gap identification system as compared to the word-word alignment system. Hence, a graph-based model (will be referred as System-I henceforth) was proposed in the baseline paper \citep{arch2016identifygaps}, which aimed at aligning the answers in semantic space through optimal undirected graph alignment techniques.\\
Similarly, directed graph-alignment approach was proposed to identify gaps in student answers in the proposed method in the paper, which took into account  directionality of edges in the answer graphs as well as  the similarity of the predicates /edge labels during alignment for better answer graph alignment as compared to System-I which lacked the above mentioned features. This may lead to the detection of gaps in student answers with increased confidence as compared to that detected using System-I.\\
As pointed out in \citep{nastase2015survey}, graphs indicate a
powerful representation formalism, which is evident from WordNet- a semantic network built from graph-based representations of meanings of words through their relations with other words. Also graph formalisms have been selected as an unsupervised learning approach to various Natural Language Processing problems such as language identification \citep{bhati2020unsupervised}, part-of-speech tagging \citep{das2011unsupervised}. As we intended to design an unsupervised model for automatic gap-identification, graph-alignment strategy seemed a natural choice. 

One of the challenges in FA lies in matching of phrases in a pair of answers that may differ lexically. In such cases, the FA system should be able to detect synonymy and paraphrasing for marking the correct gaps in the student answers. In some other cases, it is challenging for the FA system to effectively point out the salient points that are important for
defining the central idea as well as their absence in an input student answer. These salient points could be extracted out as gaps. There are some situations which solicit explanation over the central concepts as answer to assess the higher-order cognitive skills of the students. In such situations, the answers have larger length on an average as compared to the other kinds of answers, which makes it challenging to identify gaps in such answers. 

With reference to the above mentioned challenges posed by the FA problem, following are our contributions:
\begin{enumerate}
\item Graph theoretic modelling: The problem of identifying gaps in student answer has been modelled using various
formal settings, e.g., classification-based \citep{Nielsen09}\citep{bulgarov2018proposition}, pattern matching \citep{rodrigues2014system}\citep{sukkarieh2008leveraging}, automated reasoning \citep{makatchev2007combining}. Motivated by graph-based textual entailment methods \citep{Khot18}, we have presented a
rigorous graph theoretic formulation of the problem. \\
\item Unsupervised model for gap extraction: Supervised methods
require sizeable amount of gap annotated training data for developing a reasonably robust gap identification system. However, development of gap annotated dataset is time consuming and requires opinions of the qualified subject matter experts. In this work, we have proposed an unsupervised approach that does not require any gap annotated answers for training. \\
\item Directed graph alignment approach for FA: The proposed
method considers the directed graph representation of the
$\langle student\hspace{0.1cm}answer, model\hspace{0.1cm}answer \rangle$ pairs. These representations
have been aligned using an optimal graph alignment technique
to align node-pairs from two graphs. Finally, variations of
the directed FA model have been derived by applying several
pruning strategies that remove the weak alignments. The gaps
in the input student answer have been identified by performing
inference over the output alignments. \\
\item Dataset for gap identification: Three existing widely known
benchmark dataset, namely, UNT \citep{mohler2009text}, Beetle \citep{dzikovska2011beetle}, and SciEntsBank \citep{dzikovska2012towards} datasets from the short answer grading domain have been adapted to develop a dataset for validating the proposed
unsupervised gap identification model.
\end{enumerate}
\section{Related Work}
The problem addressed in this paper refers to providing formative feedback through identification of gaps in student answers on comparison with one or more model answers. 
This problem is directly related to previous works providing one of the kinds of feedback to students and indirectly related to prior works discussing other kinds of feedback for students. The details of these previous works, along with their categorization basis the kind of feedback provided to the students, are given below. 
\begin{itemize}
\item \textbf{Type 1:} For Missing or incomplete topics in student answer, a comprehensive reference solution provided by the teacher
\item \textbf{Type 2:} Score for student answer
\item \textbf{Type 3:} Entailment between student answer and reference answer
\item \textbf{Type 4:} Personalized hints to students 
\end{itemize}
An overview table, as shown below in Table \ref{overview_approach}, shows the various approaches categorized according to the types of formative feedback clearly defined above. 
\begin{table}[h!]
\scriptsize
\centering
\caption{Categorization of Various Approaches according to types of Formative Feedback}
\begin{tabular}{@{\extracolsep{\fill}}cc}
\hline
Types of Formative Feedback
& Prior Approaches / Works 
\\ \hline
\begin{tabular}[c]{@{}c@{}}Type 1 \end{tabular}
& \begin{tabular}[c]{@{}c@{}}AssiStudy 
\citep{rodrigues2014system} \end{tabular} \\ \hline
\begin{tabular}[c]{@{}c@{}}Type 2 \end{tabular}
& \begin{tabular}[c]{@{}c@{}}AssiStudy \citep{rodrigues2014system} \end{tabular} \\
& \begin{tabular}[c]{@{}c@{}}Supervised machine learning method using Graph-based features \citep{yang2021automated} \end{tabular} \\ 
&  \begin{tabular}[c]{@{}c@{}} Joint multi-
domain deep learning architecture \citep{saha2019joint}\end{tabular} \\
\hline
\begin{tabular}[c]{@{}c@{}}Type 3 \end{tabular}
& \begin{tabular}[c]{@{}c@{}}Why2-Atlas  \citep{makatchev2005analyzing}
\end{tabular} \\
&  \begin{tabular}[c]{@{}c@{}} C-rater  \citep{sukkarieh2008leveraging}  \citep{sukkarieh2009c}\end{tabular} \\
&  \begin{tabular}[c]{@{}c@{}}Textual Entailment Recognition \citep{Nielsen09} \end{tabular} \\
&  \begin{tabular}[c]{@{}c@{}}Minimal Meaningful Propositions \\ (MMP) based Textual Entailment Recognition \citep{bulgarov2018proposition}\end{tabular}  \\
&  \begin{tabular}[c]{@{}c@{}}AI-based Formative Assessment system using fast-text word-embedding approach \citep{vittorini2020ai}\end{tabular} \\
&  \begin{tabular}[c]{@{}c@{}} Deep-learning based Textual Entailment method \citep{kapanipathi2020infusing} \citep{belay2021cross} \end{tabular} \\ \hline
\begin{tabular}[c]{@{}c@{}}Type 4 \end{tabular} & 
\begin{tabular}[c]{@{}c@{}}Interactive systems OpenMark  \citep{butcher2008online} \end{tabular}\\ 
&  \begin{tabular}[c]{@{}c@{}}Intelligent tutoring systems KORBIT \citep{kochmar2021automated}\end{tabular}  \\
&  \begin{tabular}[c]{@{}c@{}}SDMentor Surgical Decision-Making Mentor \citep{vannaprathip2021intelligent} \end{tabular}
\\ \hline
\end{tabular}
\label{overview_approach}
\end{table}

As shown in Table \ref{overview_approach}, one of the prior works namely, AssiStudy \citep{rodrigues2014system} is grouped under both Type 1, Type 2 categories, while \citep{yang2021automated} is grouped under Type 2 category. 

AssiStudy relies on the similarity between the canonical forms of Reference Answer (RA) and Student Answer
(SA) computed based on their semantic similarity and lexical
matching. The feedback module in AssiStudy involves searching for missing or incomplete topics of SA in RA along with
the detailed explanations. The system provides feedback on
the student answers for enumeration and specific knowledge
kind of questions (concept-completion and definition type of
answers) in the form of answer score and proper reference
answers elaborated by the teacher which are simply comprehensive reference solutions provided by the teacher. 
The objective of our work is
very much close to feedback module of AssiStudy. However,
due to lack of implementation detail and formal evaluation
in AssiStudy, comparison with the proposed system seems
infeasible.

In \citep{zanzotto2020kermit}, heat parse trees introduced by KERMIT (Kernel inspired Encoder with Recursive Mechanism for Interpretable Trees) encode various relations in sentences and are useful universal syntactic representations of sentences. These parse trees enhance the quality of sentence embeddings generated by BERT and XLNet Transformer architectures by providing additional syntactical information.
On one hand, such parse trees could act as appropriate graphical representations for a pair of student answer and model answer and matching of such representations could  effectively determine formative feedback for the student answer. \\
On the other hand, similar such enhanced embeddings, as mentioned above, representing student answer and model answer in a pair could be generated and matching of the respective answer embeddings would generate a score indicating the extent of match between the pair of answers. The student answer could be scored accordingly.

In \citep{yang2021automated}, students' essays based on various concepts and co-occurrence relationship between the concepts are converted to concept maps. The number of nodes on the concept map for an essay indicate the number of unique concepts in the essay. The strength of the semantic and/or syntactic links between the concepts is represented by the edges of the map. \\
For each student essay, graph-based features based on a combination of structure of concept map such as global convergence, local convergence, distance between nodes, similarity to the concept maps of high-scoring students' essays and the word embeddings representation are extracted for the concept maps of students' essays. These graph-based features aim to capture the relationships between the concepts in the essay. These features for the training set of essays are used to train a multiple linear regression model. The trained model generates the score of the students' essays in the test set automatically. A BERT model pre-trained and fine-tuned on the students' essays is used to evaluate the performance of the above graph-based features in prediction of scores of students' essays. 

In \citep{saha2019joint}, a neural-based domain adaptation approach is proposed for scoring student answers. This approach is named as JMD-ASAG (Joint multi-domain deep learning architecture for automatic short answer grading). It makes use of multi-domain information to jointly learn domain specific classifiers and a generic classifier based on deep learning architectures for grading of student answers automatically. The grade/score hence obtained serves as a feedback to the students that motivates them to concentrate on specific topics where less scores are obtained and perform better in further tests. 

Type-3 category comprises of various previous works such as Why2-Atlas tutoring system \citep{makatchev2005analyzing}, C-rater \citep{sukkarieh2008leveraging}\citep{sukkarieh2009c} , and those mentioned in \citep{Nielsen09}, \citep{bulgarov2018proposition}, \citep{vittorini2020ai}.  

The formative feedback generation module in Why2-Atlas tutoring system has been modelled as an assumption-based truth maintenance system (ATMS). The ATMS is used to compute the appropriate reasoning for specific physics problems in mechanics. The deductive closures of a set of physics rules and a set of propositions representing the specific problem in mechanics domain is contained in the ATMS for the concerned specific problem, which is analogous to an ideal proof provided by the expert. 
The student utterances corresponding to the problem are converted to their formal representations which are compared to the ATMS nodes with the help of the largest common subgraph-based graph-matching algorithm. This is done to identify whether the expert provided propositions are mentioned in the student utterances. There can be direct matches of the graph nodes in which further analysis would give an idea whether the student utterance represented a particular physics statement otherwise analysis of a neighborhood of the matched nodes obtained by matching with stored physics statements would give an idea of how far the student utterance is from the desired physics statement. A student utterance is classified into semantic classes based on concepts and facts in physics problems using k-nearest neighbors algorithm. Various metrics such as those based on average precision and average recall are used to evaluate this ATMS based diagnosis system performance taking into account around 135 student utterances to two physics problems.

C-rater has adopted a concept-based scoring
scheme to enable individualized formative feedback for each student answer (concept-completion answers have been tested). At first, a sample of students’ answers (manually annotated with the evidence for each of the concepts in the test question, such that concept C entails evidence E) and the corresponding test question are made available to the system by the instructors. Then a set of model sentences are manually written by referring to the manually annotated students’ answers. The $\langle student\hspace{0.1cm}answer, model\hspace{0.1cm}answer\rangle$ pairs
are linguistically processed using a deeper parser followed by the extraction of the linguistic features using a set of hand-written rules. The extracted features are used to train a Goldmap matching model that aims at predicting entailment relation between the student answer and the corresponding model answer.

The problem of gap identification has been modelled as
a Textual Entailment Recognition (RTE) task in~\citep{Nielsen09}.
A fine
grained analysis of the student answers has been performed
by fragmenting the reference answers into facets. The analysis
of the student answers is projected as finding the entailment
relations between these facets and the student answers.

A finer-grained analysis of the student answers is carried
out in \citep{bulgarov2018proposition} where reference answer for a question is split into
its constituent propositions, known as, Minimal Meaningful
Propositions (MMPs) and the entailment relations between
a student response and each MMP in the corresponding
reference answer are determined using deep neural networks. Each of the reference answer MMPs are generally sentences extracted from the reference answer, which may contain numerous multiple sentences. 
The prediction whether the reference answer MMP (MMP is generally a single sentence) is “understood” or “not understood” by the student, is performed
in two ways: 1) Neural networks trained with GloVe word
embeddings 2) SVM model trained with hand-crafted features
comprised of the “General features” and the “Facet features”.
The general features indicate the general relations between
the reference answer MMP and student response such as
overlapping content, BLEU scores, etc. The facet features
determine lexical similarity using lexical entailment probabilities indicating entailment between the pair of MMP and
student response, syntactic information such as part-of-speech,
relevant dependency or facet relation types and dependency
path edit distances. The approach adopted in \citep{Nielsen09} generates
feedback in a fine-grained level whereas the MMPs in \citep{bulgarov2018proposition} are
defined at the sentence level, as mentioned above. Our work is slightly similar to the MMP-based method as described above in the sense that instead of applying a Machine-learning based approach to determine whether each reference MMP is addressed by the student response or not, our work involves extraction of gaps (similar to MMP but word/phrase/sentence instead of just sentence) in student answer, on comparison with model answer. In other words, in our work, the granularity of the gaps are kept flexible allowing the gaps to be annotated at
the word, phrase, clause or sentence level.

Deep-learning based textual-entailment methods such as those discussed in \citep{kapanipathi2020infusing} and \citep{belay2021cross} classify whether a student response/hypothesis text is complete with respect to the model response / premise text. In \citep{vittorini2020ai}, feedback indicating the completeness of the student solution with respect to the correct / model solution has been achieved using a fast-text word-embedding approach that has paved future possibilities of BERT embeddings, too. \\
In \citep{kapanipathi2020infusing}, a text-based entailment model is augmented with external knowledge obtained from knowledge graphs (KGs) for better entailment prediction results. Pre-trained BERT embeddings for each of the words appearing in a sequence in the premise and hypothesis texts obtained from textual entailment datasets are fed to a neural network-based Natural Language Inference (NLI) models to generate a fixed size text representation for the pair of premise and hypothesis texts. 
An external knowledge graph with the help of Personalized PageRank algorithm as discussed in \citep{page1999pagerank}, is used to obtain the most relevant neighbor nodes in the contextual subgraphs relevant to words / concepts mentioned in the premise and hypothesis texts, respectively. These contextual subgraphs are encoded using Relational Graph Convolutional Networks (R-GCN) to generate fixed graph representations for the premise-hypothesis pair. These graph representations along with the text representations mentioned earlier for the premise and hypothesis texts are fed to a feedforward classifier to predict entailment / contradiction / neutral. The premise and hypothesis text-pair can correspond to a pair of model answer-student answer respectively. The entailment results would indicate whether the model answer entails with respect to the student answer or in other words, the student answer refers to the model answer completely. The other labels, contradiction and neutral could refer to incomplete entailment of the model answer with respect to the student answer. 

In \citep{belay2021cross}, the preprocessed English premise text is translated into Amharic language using transformer and Bi-LSTM hybrid model. FastText word-embedding techniques are used to construct vectors for each of the words in the translated premise text as well as that for the preprocessed hypothesis text. Sentence-embedding methods build vector representations for the entire text (premise / hypothesis) from the vector representations of the words in each of the texts.  \\
XlNet and Bi-LSTM based deep learning models are trained with the sentence embeddings of the premise and hypothesis texts in the training data so as to contribute towards building features representing semantic information in texts from test set.  Also, dual dictionary approaches involving similarity score computation between pre-processed premise and hypothesis vectors of test set are used to obtain lexical information in the respective texts. The semantic and lexical information together are used to train multilayer perceptron (MLP) to predict the inferential relationship between premise and hypothesis text pair, such as various kinds of entailment such as forward entailment, backward entailment, bidirectional entailment, neutral, contradiction \citep{belay2021cross}. In other words, if the premise and hypothesis text pair represent a pair of model answer and student answer, respectively, the trained MLP with the semantic and lexical information can be used to predict various kinds of entailment of the model answer with respect to the student answer, indicating whether the student answer refers to the model answer completely or not.  

An AI-based formative assessment system \citep{vittorini2020ai} provides feedback to short answers or comments which form a part of each of the R code snippets of around 36 students enrolled in Health Informatics subject in the degree course of Medicine and Surgery of the University of L’Aquila (Italy). 
Feedback for the student code assignment solutions is  obtained by indicating the number of commands missing in student solution; but present in correct solution, the number of commands showing a different output, determining whether a specific comment used for explaining the result/output is correct or not.    \\
A fast-text word-embedding approach with the help of a pre-trained Italian language model trained on Common Crawl and Wikipedia is used to build sentence embeddings for the pair of student comments and correct comments. In addition to the sentence embeddings, seven distance-based features that determine the lexical and semantic similarity between a pair of student comments and correct comments are also extracted. These seven distance-based features are namely, Cosine of sentence embeddings, Cosine of (lemmatized) sentence embeddings, Word mover’s distance (WMD), Word mover’s distance (lemmatized), features based on overlap of tokens/lemmas, presence of negation in the comments are extracted for the pair of student and correct comments. An SVM classifier tuned and trained on all the above features  provided feedback to the student comment by labelling it as correct, partially correct or incorrect, with respect to the correct comment. Features were also obtained using BERT models which analyze words in comments by looking at the surrounding words (preceding or succeeding them). The classifier was trained with these BERT features. However, apparently due to less training data, convergence of the model was not achieved. This paved future possibilities of using BERT transformer approach for building improved AI-based formative assessment systems with availability of more training data.

Recent interactive systems such as OpenMark \citep{butcher2008online} and Intelligent tutoring systems such as KORBIT \footnote{\url{https://www.korbit.ai}} discussed in \citep{kochmar2021automated} and SDMentor (Surgical Decision-making Mentor) in \citep{vannaprathip2021intelligent} fall under the Type-4 category of previous formative assessment works. \\
OpenMark uses Intelligent Assessment Technologies’ software\footnote{\url{http://www.intelligentassessment.com/}} for matching of the student responses with the reference answers in their database for the generation of appropriate feedback. OpenMark is designed such that feedback at multiple levels of learning could be provided. As mentioned earlier, it is an interactive system encouraging the students to provide their responses and act on the immediate feedback provided to them, while the problem is still fresh in their minds. The students using OpenMark can provide their responses in multiple attempts. Attempts have been to the greatest extent in OpenMark regarding making the performance assessments of students as engaging as possible utilizing the modern multimedia computers to the fullest capacity. OpenMark assessments are user-friendly and flexible in the sense that students can complete them according to their schedule, stopping at any point and resuming from any point. 


KORBIT generates feedback automatically in the form of personalized hints and specifically shows how personalization of feedback can lead to improvements in student performance outcomes.\\ 
Personalized hints are generated by linguistic analysis of the question to automatically identify keywords and keyphrases in it, followed by dependency parsing on reference solutions to the question to remove parts of the solutions containing the keywords/keyphrases from question, and use of discourse modifiers to generate a grammatically correct hint from the clauses extracted from the reference solutions in the previous step. \\
The generation of personalization hints is followed by the process of evaluation of hints according to their quality and relevance for each student. 
A machine learning model based on Random Forest (RF) is trained with various sets of features depending on the complexity of the feedback generation model such as linguistic features that determine the quality of the hint from the perspective of linguistics, performance based features that consider the past performance of students such as number of attempts taken to answer the question, features considering a sizeable number of student's utterances just before the hint;
this RF based feedback selection model predicts whether a student with specific performance characteristics and if provided with a specific hint will solve the exercises in the next attempt. The student success rate regarding finding solutions to the exercises is obtained by computing the proportion of instances involving arrival at a correct solution with the help of a personalized hint. It is observed that use of the feedback selection models lead to improved student performance outcomes.

SDMentor (Surgical Decision-making Mentor) focuses on teaching surgical decision making for the preoperative and intraoperative stages of root canal treatment. \\
An observational study of teaching sessions is carried out by experienced human tutors in a dental clinic in order to get an idea of how surgical instructors  coach the students with respect to decision-making skills in the operating room, different strategies of tutoring, the principle behind the strategies, etc. The outcomes of this observational study prove very important towards building SDMentor for surgical training. \\
SDMentor constitutes surgical actions and the respective procedure with the help of a variant of the planning domain definition language (PDDL).
Based on the teaching strategy identified and hence the template corresponding to that strategy, SDMentor produces feedback with the help of the PDDL-based knowledge representation. The feedback is in the form of hints or questions posed to the students so as to enable them to find the final answers themselves. 

In the proposed FA model, as mentioned before, formative feedback is provided in the form of finding gaps in student answers which is very close to the kind of feedback provided by Type-I category of previous works, in which missing or incomplete topics of student answer in reference answer are identified, as mentioned above. Gaps in student answers have been identified earlier using a graph alignment approach, called System-I henceforth discussed in our previous work in \citep{arch2016identifygaps}. System-I aims to determine the missing entities in the student answer on comparison with the model answer through matching of the corresponding student answer graph and model answer graph with the help of a graph-alignment algorithm. 
This system is used as the baseline system for the present work and described in detail in Section \ref{baseline}. We have discussed the limitations of System-I that motivate the proposed FA model in Section \ref{motivation}.

 \section{Baseline System} \label{baseline}
\begin{figure}[h]
\centering
\includegraphics[scale = 0.5]{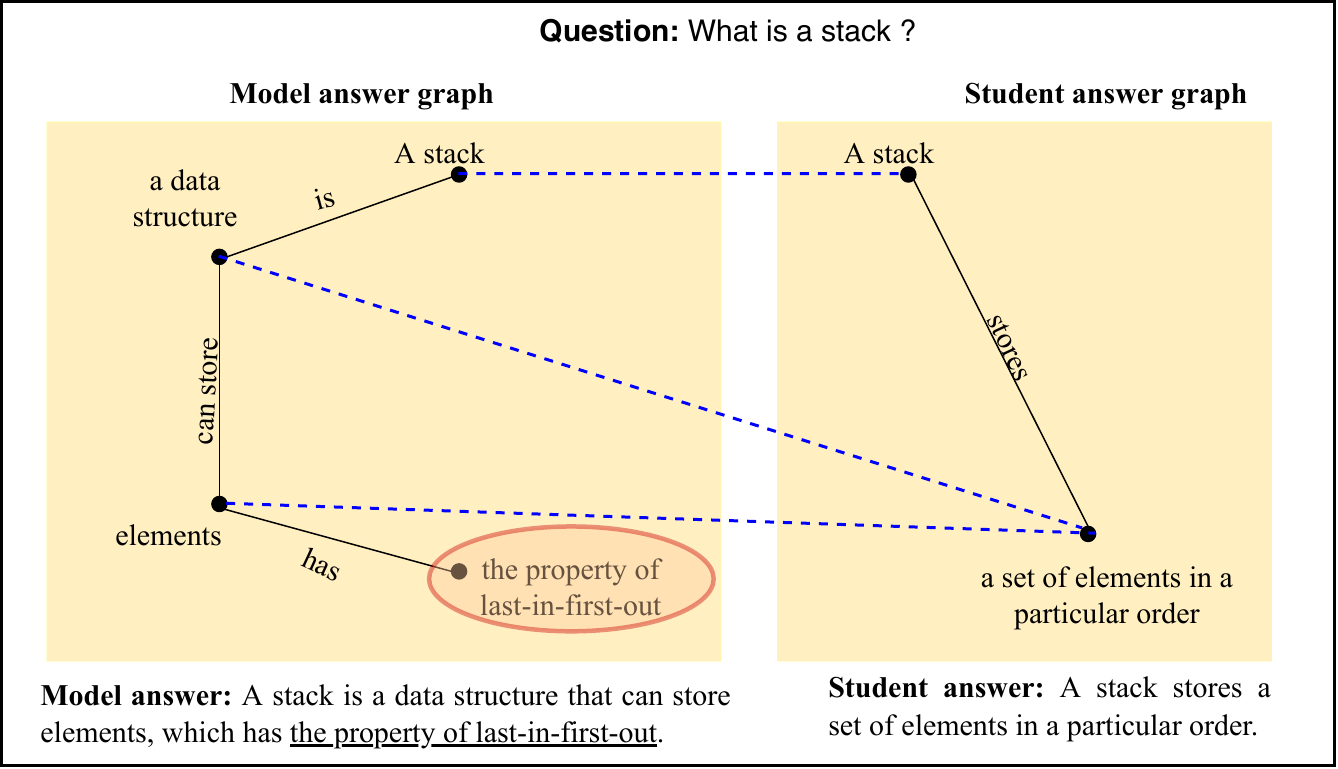}
\caption{An example for extraction of a gap in student answer using System-I. The dashed lines indicate alignment of
node-pairs from the pair of answer graphs. The elliptical portion indicates the gap detected by System-I in the student answer.}
\label{fig:system-I}
\end{figure}

As the System-I \citep{arch2016identifygaps} is the most closely related work to the present work, the System-I has been considered as the baseline system. In System-I, a
$\langle model\hspace{0.1cm}answer, student\hspace{0.1cm} answer \rangle$ pair has been represented as a pair of undirected and unweighted answer graphs constructed from the triples
of the form $\langle subject, predicate, object \rangle$ from each of the answers. The pair of answer graphs are aligned based on a global alignment approach, namely, the IsoRank algorithm
\citep{singh2008global}. The similarity between the neighborhood topologies (Structural Similarity) of a pair of nodes and the similarity between the contents of the node-pairs (Content Similarity), computed using semantic similarity measures, were considered
for determining the alignment between the pair of answer graphs. Contents of the nodes in the model answer graph that are not aligned or weakly aligned are predicted as gaps. An example of the gap identification scenario in System-I is presented in Figure\ref{fig:system-I}. It is observed that the node “the property of last-in-first-out” in the model answer graph does not find any alignment with any of the nodes in the student answer graph. Consequently, the node ``the property of last-in-first-out'' is identified as a gap in the student answer.

\section{Motivation for Proposed FA Model} \label{motivation}
The proposed work represents the answers as directed
graphs. The directionality conforms to the semantics of the
statements in the answers.

Let us consider the role of directionality in the analysis of the student answer presented in the
Example \ref{ex:4} from the SciEntsBank \citep{dzikovska2012towards} dataset. \\
For the answers presented in Example \ref{ex:4}, System-I
disregards the directionality of the edges and aligns $\langle$Oil, oil$\rangle$ and $\langle$water, Water$\rangle$ pairs. Consequently, no gap is detected as all the nodes are aligned with high similarity value. On the other hand, if the directionality of the edge ‘is less dense than’ is taken into account, the above alignments may not take place. The direction-induced alignment forces the $\langle$subject-subject$\rangle$ (Oil-Water) and $\langle$object-object$\rangle$ (water-Oil) node-pairs to align. However, this situation results in weak alignment having low similarity scores. Consequently, the appropriate gap would be detected in the student answer which seems not to be possible with the undirected graph model. This motivated us to adopt the graph representation that takes into account directionality of edges in the answer graphs for better alignment. The better graph-alignment, in turn, may lead to the detection of gaps in the student answers with increased confidence. As the predicates or the edge labels are instrumental in aligning the node-pairs, the proposed model also considers the similarity of the edge labels during alignment.
\begin{mdframed}
\begin{Example} ~\\
\label{ex:4}
\textbf{Q:} Oil, water, and corn syrup are layered in a tall, thin container. All are at the same temperature. What does the layering tell one about the density of the oil compared to water? \\
\textbf{M:} Oil is less dense than water. \\
\textbf{S:} Water is less dense than oil. 
\end{Example}
\end{mdframed} ~\\
\begin{figure*}[th]
\centering
\includegraphics[scale=0.6]{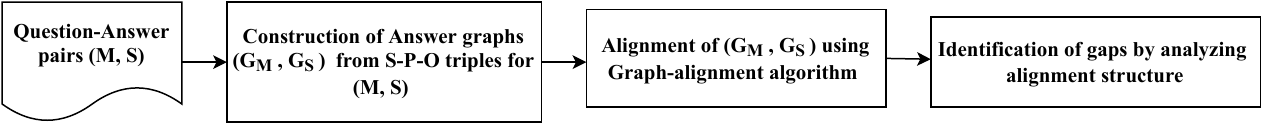}
\caption{Workflow for directed graph-alignment approach towards FA}
\label{fig:sfa_flow}
\end{figure*}
\section{Problem Definition}
The problem of automated identification of the gaps in the student answers can be formulated based on the workflow presented in Figure \ref{fig:sfa_flow}. Each answer in $\langle M, S \rangle$ pair is transformed into corresponding directed answer graphs as discussed in Section \ref{repr_answer_graph}. Subsequently, an alignment between the pair of directed answer graphs is derived by considering content of the nodes and their respective neighbours. The formal definition of the problem of aligning two answer graphs is presented in Section \ref{align_graph}. This is followed by the formal representation of the gap detection in a student answer based on the resulting alignment of the answer graphs. 

\subsection{Answer Graph Representation} \label{repr_answer_graph}
A model answer and student answer pair $\langle M, S \rangle $ is represented as a pair of directed graphs $\langle G_{M}, G_{S} \rangle $. 
$G_{M}$ and $G_{S}$ are constructed by extracting subject-predicate-object triples $\langle s, p, o \rangle $ from $M$ and $S$, respectively. 
\noindent \begin{definition}{\textit{Triple Set ($M_{t}$) for Model Answer ($M$)}} \\
A triple set ($M_{t}$) is the set of all the triples that are extracted from the model answer ($M$) and is of the form $\{\langle s, p, o \rangle\}$. $s$, $p$, $o$ are called the subject, predicate and object, respectively.
\end{definition}
\begin{equation}
    \mathcal{S}_M = \{s\mid\langle s, p, o \rangle \in M_t \} 
\end{equation}
\begin{equation}
    \mathcal{O}_M = \{o\mid\langle s, p, o \rangle \in M_t \}
\end{equation}
\begin{equation}
    \mathcal{P}_M = \{p\mid\langle s, p, o \rangle \in M_t \}
\end{equation}

$\mathcal{S}_M$ and $\mathcal{O}_M$ represent the set of subjects and objects belonging to the triples in $M_{t}$ that have been extracted from $M$ using an Information Extraction (IE) tool. It is to be noted that $\mathcal{S}_M \cap\mathcal{O}_M$ may be a non-empty set as an object in one triple can act as a subject for another triple and vice versa. $\mathcal{P}_M$ represents the set of predicates belonging to the triples in $M_{t}$ generated from $M$.
It is also to be considered that a predicate $p\in \mathcal{P}_M$ can appear in the answers in different linguistic variations. Here, it is assumed that $p$ is used in a canonical notion where different linguistic variations of a given predicate are mapped to $p$. The process for converting  the linguistic variations of a given predicate to its canonical representation is discussed in Section~\ref{sec:canonical}. Triple set ($S_{t}$), $\mathcal{S}_S$, $\mathcal{O}_S$, and $\mathcal{P}_S$ of a student answer $S$ are defined similarly.
\begin{definition}{\textit{Model Answer Graph ($G_M$)}} \\
The directed and labeled graph $G_M = \langle V_M, E_M \rangle$ is defined by a finite set of vertices $V_M = \mathcal{S}_M \cup \mathcal{O}_M$\footnote{For the sake of simplicity in notation, the same symbol is used to represent the set of subjects/objects and their corresponding node representations. Thus, $s\in \mathcal{S}_M$ or $o \in \mathcal{O}_M$ will be treated to be the same as their corresponding node representations, i.e., subject node ($N_M^{s}$) or object node ($N_M^{o}$) respectively.} and a finite set of directed edges $E_M$. The set of edges ($E_M$) includes only the predicates that are present in the corresponding model answer and has one-to-one correspondence with $\mathcal{P}_M$, i.e., the set of predicates present in $M$. The set edges $E_M$ is defined as:
\begin{equation}
\begin{split}
 E_M = \{ \langle N_M^{s} \xrightarrow{\quad p \quad}  N_M^{o}\rangle ~\mid~ N_M^{s}\in \mathcal{S}_M~ \wedge ~ \\ N_M^{o} \in \mathcal{O}_M ~\wedge 
     p \in \mathcal{P}_M \wedge  \langle N_M^{s}, p, N_M^{o}\rangle \in M_t \} 
\end{split}
\end{equation}
\end{definition}
Student answer graph ($G_S = \langle V_S, E_S \rangle$) can be defined similarly. The implementation details of this part of the problem is presented in Section~\ref{sec:canonical}. 

\subsection{Answer Graph Alignment} \label{align_graph}
The alignment of these directed answer graphs considers the directionality and the labels (in canonical form) of the edges. 
\begin{definition}{\textit{Aggregated Predicate Set}} \\
The aggregated predicate set ($\mathcal{P}$) is defined as:
\begin{equation}
    \mathcal{P} = \mathcal{P}_M \cup \mathcal{P}_S 
\end{equation}
where $\mathcal{P}_M$ is the set of all the predicates in $M_t$ and $\mathcal{P}_{S}$ is the set of all the predicates in $S_t$. 
\end{definition}

\begin{definition}{\textit{Predicate Induced Alignment}} \\
Given two answer graphs $G_M$ and $G_S$, the predicate induced alignment for a predicate $p \in \mathcal{P}$ is defined as:
\begin{equation}
\label{eq:pia}
\begin{split}
    \mathcal{A}_{p} = \{ (\langle N_M^{s}, N_S^{s}\rangle, \langle N_M^{o}, N_S^{o}\rangle)~\mid~ \exists_{N_M^{s}, N_M^{o}} \langle N_M^{s}, p, N_M^{o}\rangle \in M_t \\ \wedge~ \exists_{N_S^{s}, N_S^{o}} \langle N_S^{s}, p, N_S^{o}\rangle \in S_t \} 
\end{split}
\end{equation}
\end{definition}
It is to be noted that a triple might not have an object in general case. In such cases, a dummy node is used to represent the null object node. Such dummy nodes are ignored during alignment.
\begin{definition}{\textit{Predicate Induced Alignment Score}} \\
Let $\sigma(x,y)$ denotes the similarity of contents in the node-pair $\langle x, y\rangle $. More information about $\sigma(x,y)$ is presented in Section~\ref{sec:alignment} (Equation~\ref{eqn:sim}). The predicate induced alignment score for an alignment $a_p \in \mathcal{A}_{p}$ is:
\begin{equation}
 \xi(a_p) = \sigma(N_M^{s}, N_S^{s}) + \sigma(N_M^{o}, N_S^{o})   
\end{equation}
where, $a_p = (\langle N_M^{s}, N_S^{s} \rangle, \langle N_M^{o}, N_S^{o}\rangle)$
\end{definition}
Let $\mathcal{L} = \mathcal{A}_{p_1} \times \mathcal{A}_{p_2} \times .... \times \mathcal{A}_{p_n} $ be the Cartesian product of the predicate induced alignment sets (i.e., $\mathcal{A}_{p}$). The set $\mathcal{L}$ represents all possible combinations of the predicate induced alignments over the $n$ predicates $p_1, p_2, \ldots p_n$. A member $\mathcal{L}_k\in \mathcal{L}$ is one possible combination having exactly one member from each predicate induced alignment set ($\mathcal{A}_{p_i}$). Let $\mathcal{L}_k = \{a_{p_1}^{(k)}, a_{p_2}^{(k)}, \ldots, a_{p_n}^{(k)}\}$ be the $k^{th}$ tuple in the set of combinations of the predicate induced alignments. Let $a_{p_i}^{(k)}$ be the alignment chosen from $\mathcal{A}_{p_i}$ for $\mathcal{L}_k$. Each such $\mathcal{L}_k$ can be assigned an alignment score by  aggregating the score of the individual alignments $a_{p_i}^{(k)}$. The optimal alignment of directed answer graphs is defined as:
\begin{equation}
    \mathcal{L}_{*} = \argmax_{k}\sum\limits_{i=1}^{n} \xi(a_{p_i}^{(k)})
\end{equation}
The methods and the components related to the answer graph alignment problem are presented in the Section~\ref{sec:alignment}. 
\subsection{Detection of Gaps in $S$}
After the alignment between $G_M$ and $G_S$ is obtained, the next step is to detect the gaps in student answer $S$. The words or phrases in $M$ that are missing from $S$ are extracted as gaps in $S$. The set of gaps in a given student answer ($S$) can be represented as:
\begin{equation}
    G(S) = \{N_M~\mid~ \neg~\exists_{N_S}N_S \in V_S ~\land~ (N_M, N_S) \in \Phi^{*}\} 
\end{equation}
The set $\Phi^{*}$ is obtained by extracting the node level alignment pairs (subject-subject or object-object) from each entry in $\mathcal{L}_{*}$. It is to be noted that $\abs{\Phi^{*}}\leq 2\abs{\mathcal{L}_{*}}$ considering the fact that the dummy nodes are ignored.\\
This basic alignment aligns different node-pairs $(N_{M}, N_{S}) \in \mathcal{S}_M \times \mathcal{O}_M$ with a score. It may also be noted that the node-pairs having very low $\sigma$-score may be pruned to establish meaningful alignment using some filters. Consequently, the gaps can be identified by considering only remaining alignment.
\begin{equation}
    G^{F}(S) = \{N_{M}~\mid~\neg~\exists_{N_S} N_S \in V_S ~\land~ (N_M, N_S) \in \Phi_{F}^{*}\}
\end{equation}
where $\Phi_F^{*}$ is the set of aligned node- pairs after applying a filter condition $F$ (such as node-pairs having $\sigma$ value less than a threshold) on $\Phi^{*}$. 
The models related to the gap identification problem are discussed in the Section~\ref{filters} and Section~\ref{sec:gap_detection}.

\section{Proposed FA Model}
The proposed FA model adopts a graph theoretic approach
towards identifying gaps in the student answers. This section
describes the various stages of the directed graph-alignment
approach relevant to the proposed FA model. Each stage is
illustrated with a toy example showing the output of the
proposed FA model at each stage, except for Section \ref{sec:gap_detection}
that shows various cases for gap detection.

An input student answer and the corresponding model answer
are transformed into their corresponding graph representations.
The constructed graphs are aligned node-wise based on content
and structural similarity. The nodes in the model answer graph,
for which alignment with the student answer graph could not
be established, are detected as gaps in student answer.
\subsection{Relation Extraction and Construction of Answer Graphs}\label{sec:canonical}
Unstructured text answers contain several natural language
statements. The sentences that are in passive mode are transformed into their corresponding active form. Triples of the
form subject-predicate-object are extracted from sentences in
the answers using well-known Information Extraction tool
ClausIE\footnote{ClausIE-Clause-Based Open Information Extraction: \url{https://www.mpi-inf.mpg.de/departments/databases-and-information-systems/software/clausie/}} and are used to form the edges in an answer graph.
The direction of an edge is from “subject” to “object” and
predicate acts as the label of the edge. 

It may be noted that the subject-predicate-object (s-p-o) triples in answer graphs can be obtained using alternative methods involving Link Parser, Shallow Parsers such as those mentioned in \url{https://pypi.org/project/cltl.triple-extraction/}, \citep{dali2008triplet}, instead of the ClausIE tool used in the proposed approach. In other words, the design and  working of the proposed approach does not depend on the ClausIE tool. So, the ClausIE tool could be configured and replaced by other alternative methods as mentioned above, for any future enhancements in the proposed approach. 

The edge labels (i.e., the predicates) in the answer graph
may be semantically similar but lexically different. The answer
graphs thus may be represented in a compact form (called
canonical answer graphs) by treating semantically similar edge
labels to be the same.

\begin{figure*}[h!]
\centering
\includegraphics[width=1.0\textwidth, scale=1.2]{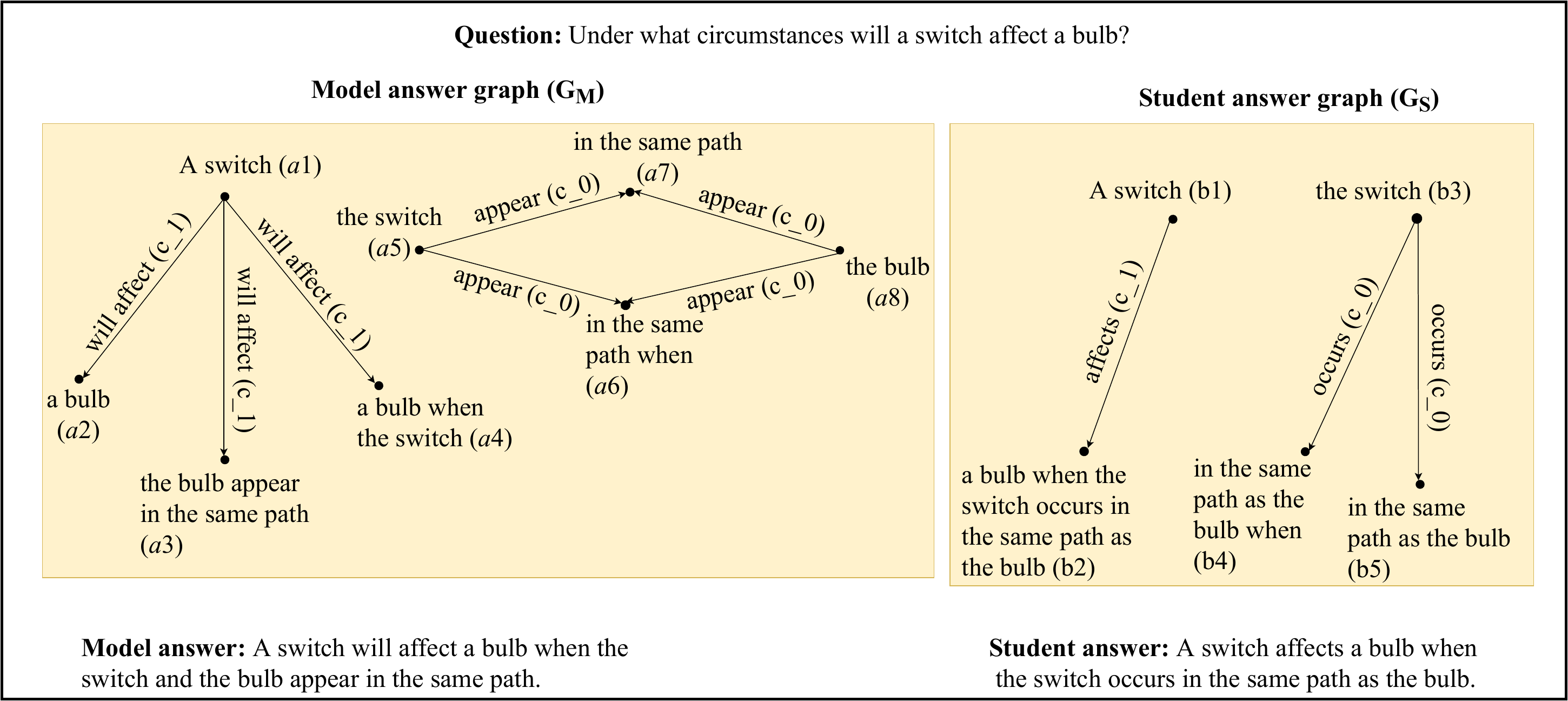}
\caption{Canonical answer graphs for a $\langle$Model Answer, Student Answer$\rangle$ pair. The graphs obtained by removing the group IDs on the edges represent the original answer graphs and those obtained by having only the group IDs represent the canonical answer graphs. The nodes are also assigned with IDs to bring in aesthetic clarity in the subsequent illustrations.}
\label{fig:new1}
\end{figure*}

The predicates in the student answer graphs and model
answer graphs are clustered on the basis of their similarity
and each cluster is assigned a unique ID. The clustering of the
predicates in the pair of answer graphs is performed offline and
only once for answer-pairs related to all questions in a dataset.
In order to create clusters of similar predicates, predicates from
all the answer graphs (for both model and student answers
in a dataset) are collected. Each predicate is transformed
into a vector representation using Continuous-Bag-of-Words
(CBOW) model. The mean of the vectors of the words in a
predicate phrase is considered as the vector representation of
the predicate. These vectors are then clustered into coherent
semantic groups with the help of K-means algorithm \citep{kanungo2002efficient}.
The number of groups k is determined by the \textit{Elbow method}
\citep{abramyan2016cluster}. All predicates belonging to the same group are assigned
a unique ID. This ID is represented as $c\_i$ where $i \in \mathbb{W}$,
the set of whole numbers. This sub-step is performed for
answer graphs corresponding to each ($M, S$) in dataset. The
edge predicates in the student answer graph and model answer
graph, corresponding to an ($M, S$) pair are replaced with
respective group IDs. Canonical student and model answer
graphs are depicted in Figure \ref{fig:new1}. The notations for the nodes and
IDs for the edge labels are mentioned in brackets alongside
the original nodes and edge labels. These notations and IDs
have been used for the further process in alignment of answer
graphs.
\subsection{Alignment of Canonical Answer Graphs}\label{sec:alignment}
To derive the alignment of canonical representations of
the pair of model and student answer graphs, the Similarity
Flooding Algorithm~\citep{park2015malware}~\citep{melnik2002similarity} is employed. This algorithm is capable of deriving inexact alignment between a pair of
directed answer graphs that may differ in size. It is an iterative algorithm based on fixpoint computation. This algorithm assumes that whenever any two nodes in a pair of graphs are found to be similar, the similarity of their adjacent nodes increases due to the flow of similarity from the matched node-pairs to their neighbors. Thus, over a number of iterations, the initial similarity of any two nodes propagates through the graphs to their neighbours. The algorithm stops after a fixpoint (or convergence point) has been reached, i.e., similarity of all node-pairs stabilize.

Initial similarity score between a node-pair is determined
by measuring similarity of 300-dimensional Word2vec vectors
\citep{mikolov2013efficient} representing the contents of the node-pairs. Let the
initial similarity score for a node-pair (x, y) be denoted as
$\sigma^{0}(x, y)$ (Equation 12). The word2vec vectors are trained
on a large Wikipedia corpus using Continuous-Bag-of-Words
(CBOW) model. In word-vector based similarity computation,
the Word2vec vectors are constructed for each of the words
in the phrases corresponding to a node-pair. The mean of the
Word2vec vectors corresponding to each node in the node-pair
is obtained. Cosine similarity between the pair of mean vectors
represents the word-vector similarity score for a node-pair.
\begin{equation}
\label{eqn:sim}
    \sigma^{0}(x, y) = \cos{(\mathbf{x}, \mathbf{y})}=\frac{\mathbf{x}.\mathbf{y}}{\sqrt{\sum_{l=1}^n \mathbf{x}_{l}^2}\sqrt{\sum_{l=1}^n \mathbf{y}_l^2}}
\end{equation}
where $\textbf{x}$ and $\textbf{y}$ represent the mean vectors corresponding to the nodes $x$ and $y$, respectively, $n$ is the length of $\textbf{x}$ and $\textbf{y}$.

The crucial three stages of aligning answer graphs using the Similarity Flooding Algorithm are presented in the following sections:

\subsubsection{Determination of Pairwise Connectivity Graph (PCG)} Let $\langle G_M, G_S \rangle$\footnote{For the sake of simplicity, we symbolize the canonical answer graphs also with $G_M$ and $G_S$.} represents a pair of canonical model and student answer graphs. A graph, called the Pairwise Connectivity Graph (PCG), is derived from $\langle G_M, G_S \rangle$. Each node in the PCG is formed using a pair of nodes; one from $G_M$ and the other from $G_S$. Such a node in PCG represents a possible alignment node-pair from $\langle G_M, G_S \rangle$. Each edge in PCG, denoted by $\langle (N_M^{s}, N_{S}^{s}), p, (N_M^{o}, N_S^{o}) \rangle$ represents an alignment between a pair of edges from $\langle G_M, G_S \rangle$ following Equation~\ref{eqn:PCGedge}.
\begin{equation}
\label{eqn:PCGedge}
\begin{split}
    \langle (N_M^{s}, N_{S}^{s}), p, (N
_M^{o}, N_S^{o}) \rangle \in PCG(G_M, G_S) \\ \iff \langle N_M^{s}, p, N_M^{o} \rangle \in E_M \wedge \langle N_{S}^{s}, p, N_{s}^{o} \rangle \in E_S
\end{split}
\end{equation}

\begin{figure}[h!]
\centering
\includegraphics[scale=0.4]{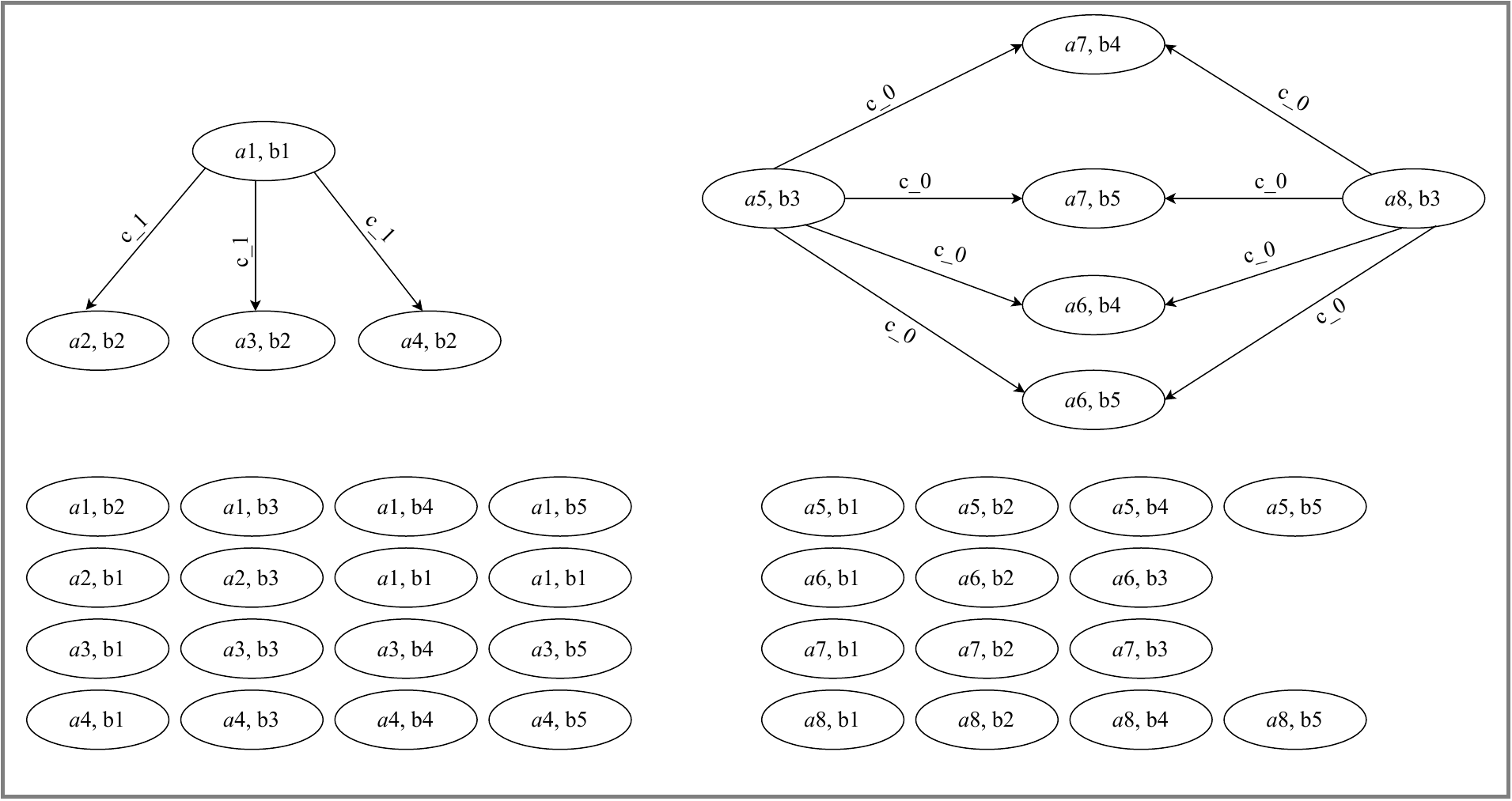}
\caption{Pairwise Connectivity Graph (PCG) from $G_M$ and $G_S$}
\label{fig:two}
\end{figure}

$PCG$ for $ \langle G_M, G_S \rangle $ from Figure \ref{fig:new1} is shown in Figure \ref{fig:two}. 
It is to be noted that PCG considers all node- pairs from $G_M$ and $G_S$. This might seem to be contradicting the predicate induced alignment presented in Equation~\ref{eq:pia}. The node- pairs that are not part of the predicate induced alignment have been accommodated to comply with the Similarity Flooding Algorithm. The similarity values for these node-pairs will diminish over the iterations as they do not have any neighbor to contribute to their initial similarity values. 

Consider node-pairs $(a5, b3)$ and $(a7, b4)$ in Figure \ref{fig:two}. They are connected by an edge labeled $c\_{0}$. The presence of the edge  $c\_0$ indicates that the similarity of $a5$ and $b3$ affects the similarity of $a7$ and $b4$. The node-pair $(a7, b4)$) is a neighbour of ($(a5, b3)$ in the constructed PCG. 
\subsubsection{Construction of Induced Propagation Graph (IPG)}
An Induced Propagation Graph (IPG) is constructed from PCG by adding a reverse edge for every edge in a PCG. The edges in the IPG are weighted. The weights, called the propagation coefficients, represent the proportion of similarity flow from a PCG node to its neighbours. The weights (ranges from 0 to 1) are determined by considering uniform similarity flow from a node to its neighbors. 

\begin{figure}[h!]
\centering
\includegraphics[scale=0.45]{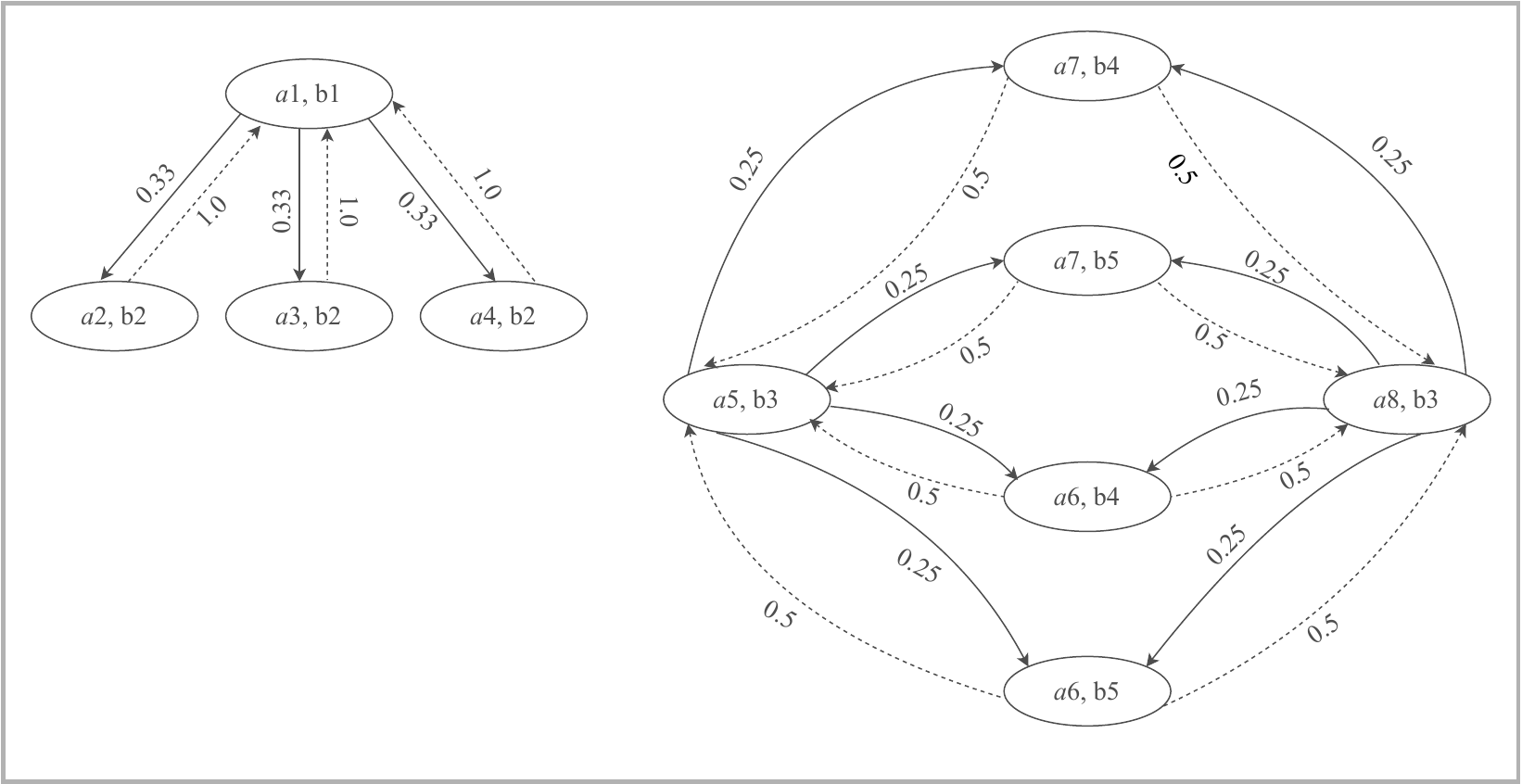}
\caption{Construction of Induced Propagation Graph from PCG. The disconnected nodes from the PCG are not shown in the figure for simplicity. }
\label{fig:three}
\end{figure}

It is observed from Figure \ref{fig:three} that the semantic similarity of the node-pair ($a5, b3$) i.e. $(the\hspace{0.1cm}switch, the\hspace{0.1cm}switch)$ is propagated to its neighbouring nodes ($a7, b4$) and ($a7, b5$) by a  weight of 0.25 (1/4). This is due to the fact that the semantic similarity of the node-pair ($a5, b3$) (i.e., 1.0) is propagated uniformly to four PCG neighbors, namely, ($a7, b4$), ($a7, b5$), ($a6, b4$) and ($a6, b5$) . As there are two similar $(c\_0)$ incoming edges to ($a7, b4$) from ($a5, b3$) and ($a8, b3$), the weights on the reverse edge from ($a7, b4$) to ($a5, b3$) and that from ($a7, b4$) to ($a8, b3$) are set as 0.5. 

\subsubsection{Fixpoint Computation of Node-Pair Similarity}
Let $\sigma(x,y)$ denote similarity between nodes $x \in V_M$ and $y \in V_S$. The Similarity Flooding Algorithm involves iterative computation of $\sigma$ values. Let $\sigma^{0}$ denote the initial alignment score between nodes of a pair of answer graphs and $\sigma^{i}(x,y)$ denote the alignment score between $x$ and $y$ after the $i^{th}$ iteration. 

The similarity value of a node-pair $\langle x, y\rangle$ in IPG in $(i+1)^{th}$ iteration, denoted as $\sigma^{i+1}$, is obtained using similarity value $\sigma^{i}$ computed in the $i^{th}$ iteration using Equation~\ref{fixpoint-eq}.
\begin{equation} \label{fixpoint-eq}
    \sigma^{i+1}(x, y) = \frac{\sigma^{i}(x, y)+f(\sigma^{i}(x, y))}{\sigma^{i+1}_{max}}
\end{equation}

where:
\begin{equation} \label{eq:max}
\sigma^{i+1}_{max} = \max\limits_{(x, y) \in V_M \times V_S}\Big(\sigma^{i}(x,y) + f(\sigma^{i}(x,y))\Big)
\end{equation}
\begin{equation} \label{neigh}
f(\sigma ^{i}(x, y)) = \\ \sum_{\substack{(a_{u},p,x)\in G_M, \\ (b_{u},p,y)\in G_S}} \sigma^{i}(a_{u}, b_{u}) \cdot w((a_{u}, b_{u}), (x, y)) 
\end{equation}

The similarity value for a node-pair $\langle x, y\rangle$ in IPG in the present iteration i.e., $(i + 1)^{th}$ iteration, denoted by $\sigma^{i+1}(x, y)$, is obtained from the sum of similarity value for a node-pair $\langle x, y\rangle$ in IPG in previous iteration i.e.,  $i^{th}$ iteration denoted by $\sigma^{i}(x, y)$ and $f(\sigma ^{i}(x, y))$. The term $f(\sigma^i(x,y))$ in Equation~\ref{fixpoint-eq} is the weighted sum of the contributions from all the node-pairs that are the neighbors of $\langle x, y\rangle$ in the IPG. It is to be noted that the weighted sum of the inverse edge $\langle x, y\rangle$ to $\langle a_u, b_u)\rangle$ is not considered in  Equation~\ref{fixpoint-eq} as the flow of similarity is towards the reverse direction. The weight of the contribution from a neighboring node-pair $\langle a_u, b_u \rangle$ to $ \langle x, y \rangle$ is represented by $w((a_u, b_u), (x,y))$ in Equation \ref{neigh}. 
Finally, the sum is normalized with $\sigma_{max}^{i+1}$.

For the IPG presented in Figure \ref{fig:three}, the unnormalized $\sigma^{1}(a7, b4)$ value after the first iteration is computed as follows:
\begin{equation}
    \sigma^{1}(a7, b4) = \sigma^{0}(a7, b4) + 0.25 \times \sigma^{0}(a5, b3) + 0.25 \times \sigma^{0}(a8, b3)
\end{equation}

The iterations are continued until the Euclidean distance of the residual vectors $\Delta (\overrightarrow{\sigma^{i+1}}, \overrightarrow{\sigma^i})$ attains a value that is less than $\epsilon$. The convergence is forced if the number of iterations exceeds maximum limit ($I_{max}$) without meeting natural convergence criteria. It is to be noted that the final similarity value of a node-pair ($\sigma^c(x,y)$) does not reflect the absolute similarity between $x$ and $y$; rather it is a relative similarity value where at least one of the node-pairs in the IPG graph has $\sigma^c$ value equal to 1.0.

The maximum of final alignment score ($\sigma^{c}$) and initial alignment score ($\sigma^{0}$) for each node-pair ($x, y$) is computed and denoted by $\sigma_{f}$ as shown below:
\begin{equation} \label{align-max}
\sigma_f = \max(\sigma^c(x, y), \sigma^0(x, y))
\end{equation}

The node-pairs are arranged in a list in the decreasing order of $\sigma_{f}$ scores which are in the range of 0 to 1 as shown in Table \ref{fixpoint}.
It is observed from Table \ref{fixpoint} that the final alignment scores ($\sigma^{c}$) might become smaller than initial alignment scores ($\sigma^{0}$).
This is due to the reason that the final alignment score for a node-pair in a particular iteration is obtained using normalized value of sum of alignment score in the previous iteration and alignment scores from the neighbouring node-pairs as shown in Equation \ref{fixpoint-eq}. The final alignment score of a node-pair is obtained by above normalization in every iteration.  

\begin{table}[h!]
\centering
\caption{Fixpoint values for alignment between $G_M$ and $G_S$ as in Figure \ref{fig:new1}}
\begin{tabular}{@{\extracolsep{\fill}}cccc}
\hline
Node from $G_M$ & Node from $G_S$ & $\sigma^{0}$ & $\sigma^{c}$ \\ \hline
A switch \textit{(a1)} & The switch \textit{(b3)} & 1.0 & 1.0 \\ \hline
\begin{tabular}[c]{@{}c@{}}a bulb when \\ the switch\end{tabular}  \textit{(a4)} & \begin{tabular}[c]{@{}c@{}} a bulb when the \\ switch occurs in the \\ same path as the \\ bulb \textit{(b2)}\end{tabular} & 0.881 & 0.333 \\ \hline
\begin{tabular}[c]{@{}c@{}}the bulb appear in \\ the same path  \textit{(a3)}\end{tabular} & \begin{tabular}[c]{@{}c@{}}in the same path \\ as the bulb when\textit{(b4)}\end{tabular} & 0.859 & 0.167 \\ \hline
\begin{tabular}[c]{@{}c@{}}the bulb appear in \\ the same path \textit{(a3)}\end{tabular} & \begin{tabular}[c]{@{}c@{}} a bulb when the \\ switch occurs in the \\ same path as the \\ bulb \textit{(b2)} \end{tabular} & 0.813 & 0.211 \\ \hline
\begin{tabular}[c]{@{}c@{}}the bulb appear in \\ the same path  \textit{(a3)}\end{tabular} & \begin{tabular}[c]{@{}c@{}}in the same path \\ as the bulb\textit{(b5)}\end{tabular} & 0.761 & 0.165 \\ \hline
\begin{tabular}[c]{@{}c@{}}the switch \textit{(a5)}\end{tabular} & \begin{tabular}[c]{@{}c@{}}in the same path \\ as the bulb when \textit{(b4)}\end{tabular} & 0.383 & 0.0005  \\ \hline
\end{tabular}
\label{fixpoint}
\end{table}

\subsection{Filters for Extraction of Best Matching Node-Pairs}
\label{filters}
This section depicts the output of the proposed FA model at the stage where various filters are used to extract the best matching node-pairs after alignment of directed answer graphs.
\subsubsection{Threshold Filter} With the help of a threshold filter, the node-pairs having similarity value below a particular threshold are pruned. In the proposed system, a threshold value of 0.5 has been taken. The remaining subset of node-pairs after pruning represents the best matches for a pair of answer graphs. It is to be noted that a many-to-many alignment is possible if the node-pairs are extracted using the threshold filter. Let $\mathcal{T}$ denote the set of node-pairs that are obtained using the threshold filter for a pair of answer graphs $\langle G_M, G_S \rangle$. Let $\sigma^{0}$ denotes the initial alignment score of a node-pair ($N_M, N_S$), where $N_M \in V_M$ and $N_S \in V_S$ and $\sigma^{c}$ denotes the final alignment score of the node-pair ($N_M, N_S$) after a number of iterations of the graph-alignment algorithm.

 \begin{equation}
 \begin{split}
       \mathcal{T} = \{\langle (N_M, N_S), \sigma^f\rangle~\mid~(\sigma^{c}(N_M, N_S) > \tau) ~ \lor ~  (\sigma^{0}(N_M, N_S) > \tau) \} 
\end{split}
\end{equation}
where $\tau$ is predefined threshold value and $\sigma_f$ is computed using the Equation \ref{align-max}. Henceforth, the gap detection model using the threshold filter will be denoted as \textbf{FA-T}.

\subsubsection{Exact filter} While the threshold filter may output many-to-many alignments, the exact filter ensures a one-to-one alignment between $V_M$ and $V_S$. A greedy selection strategy is used to align a node from $V_M$ to a node in $V_S$. The node-pairs are obtained by imposing one-to-one alignment constraint over $\mathcal{T}$, (i.e., the threshold filter pairs). Let us consider that the predicate (following the First-Order-Logic syntax) \emph{Connect (a, b)} represent that a node $a$ is aligned to another node $b$ under an alignment. To identify the $i^{th}$ node in a model answer $M$, the symbol $N_{M,i}$ is used. The $l^{th}$ node in a student answer graph is represented using the symbol $N_{S,l}$.
\begin{equation}
\begin{split}
   \forall_{i,j,l}~Connect(N_{M, i}, N_{S, l})~\wedge~ Connect(N_{M, j}, N_{S, l}) \implies N_{M, i} = N_{M, j}  \label{eq:connect1} 
\end{split}
\end{equation}
The set of resulting matches using exact filter,  $\mathcal{E}$,  is defined as:
\begin{equation}
    \begin{split}
        \mathcal{E} = \{\langle (N_M, N_S), \sigma^f \rangle \in \mathcal{T}~\mid~ Equation~\ref{eq:connect1}~is~satisfied\}
    \end{split}
\end{equation}
In the subsequent part, \textbf{FA-E} will be used to denote the gap extraction model using the exact filter.
\subsubsection{Best filter} In this filter, a one-to-one alignment is obtained between the nodes of $G_M$ and $G_S$ such that the cumulative similarity value of all the aligned node-pairs is maximum. A one-to-one alignment $\Phi_{k}$ between $G_{M}$ and $G_S$ is obtained such that each node of $G_{S}$ is aligned to at most one node of $G_M$.
An optimal one-to-one alignment $\Phi^{*}$ is defined as a one-to-one alignment $\Phi_{k}$ having the maximum alignment score. The optimal alignment $\Phi_{k}^{*}$  is obtained as follows:
\begin{equation}
  \displaystyle \Phi^{\star} = \argmax_{k}{\sum_{N_M \in V_M}\sum_{N_S \in V_S}\sigma(N_{M}, N_{S}) \bullet \mathtt{Align}(\Phi_{k}, N_{M}, N_{S})}
\end{equation}
\begin{equation*}
     \mathtt{Align}(\Phi_{k}, N_{M}, N_{S}) = \begin{cases}
                        1 & \text{if $N_{M}, N_{S}$ are aligned} \\ & \text{under } k^{th} \text{ one-to-one alignment} \\
                        0 & \text{otherwise}
                        \end{cases}
\end{equation*}
The problem of finding optimal alignment can be modelled as a stable-marriage problem. The optimal one-to-one alignment for the \emph{best filter} is obtained using the Hungarian matching algorithm \citep{kuhn1955hungarian}. Let the set of aligned node-pairs in $\Phi_{k}^{*}$ be denoted as $\mathcal{B}$. The gap extraction model using the best filter is denoted by \textbf{FA-B}.

On referring the fixpoint values of node-pairs in Table \ref{fixpoint}, the node-pairs obtained in the final alignment between $G_M$ and $G_S$ using the threshold filter, exact filter and best filter are shown in Table \ref{fil}.

\begin{table}[h!]
\centering
\caption{Pruned alignment between $G_M$ and $G_S$ using different filters.  $\sigma_{f}$ values are mentioned in case of pruned alignment using \textit{threshold filter} and \textit{exact filter}. A threshold value of 0.5 has been considered. $\sigma^{c}$ values (final alignment scores) are mentioned in case of optimal alignment using \textit{best filter}. Thereafter for gap identification, $\sigma$ notation is used to indicate either of $\sigma_{f}$ or $\sigma^{c}$ which is understood according to the type of filter used in proposed FA model.}
\begin{tabular}{@{\extracolsep{\fill}}ccccc}
\hline
Node in M & Node in S & \multicolumn{3}{c}{Similarity Score} \\ \hline
 &  & \begin{tabular}[c]{@{}c@{}}threshold \\($\sigma_{f}$)\end{tabular} & \begin{tabular}[c]{@{}c@{}}exact \\ ($\sigma_{f}$) \end{tabular} & \begin{tabular}[c]{@{}c@{}} best \\ ($\sigma^{c}$) \end{tabular}  \\ \hline
A switch (a1) & The switch (b3) & 1.0 & 1.0 & 1.0 \\ \hline
\begin{tabular}[c]{@{}c@{}}a bulb when the\\ switch (a4)\end{tabular} & \multicolumn{1}{l}{\begin{tabular}[c]{@{}l@{}}a bulb when the\\ switch occurs in \\the same path as \\ the bulb (b2)\end{tabular}} & 0.881 & 0.881 & 0.333 \\ \hline
\begin{tabular}[c]{@{}c@{}}the bulb appear in\\ the same path (a3)\end{tabular} & \begin{tabular}[c]{@{}c@{}}in the same path as\\ the bulb when (b4)\end{tabular} & 0.859 & 0.859 & 0.165 \\ \hline
\begin{tabular}[c]{@{}c@{}}the bulb appear in\\ the same path (a3)\end{tabular} & \multicolumn{1}{l}{\begin{tabular}[c]{@{}l@{}}a bulb when the\\ switch occurs in \\ the same path as \\ the bulb (b2)\end{tabular}} & 0.813 & - & - \\ \hline
\begin{tabular}[c]{@{}c@{}}the bulb appear in\\ the same path (a3)\end{tabular} & \begin{tabular}[c]{@{}c@{}}in the same path\\ as the bulb (b5)\end{tabular} & 0.761 & - & - \\ \hline
\end{tabular}
\label{fil}
\end{table}

\subsection{Identification of Gaps in Student Answer $S$}
\label{sec:gap_detection}
\begin{figure*}[h!]
  \centering
  \subcaptionbox{Gap identification for Case 1\label{fig:case1}}{%
  \includegraphics[scale=0.4]{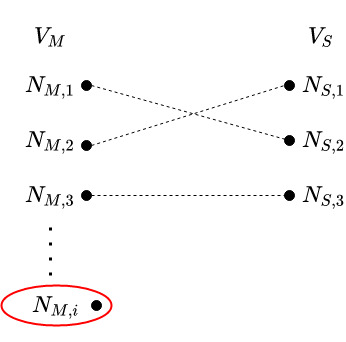}}\quad
  \subcaptionbox{Gap identification for Case 2\label{fig:case2}}{%
  \includegraphics[scale=0.4]{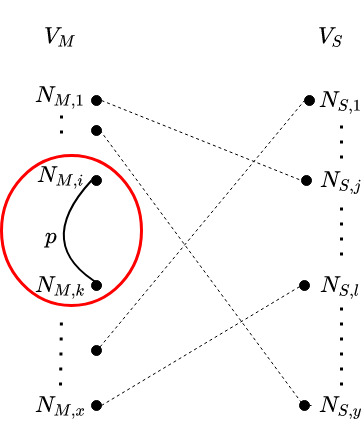}}\quad
  \subcaptionbox{Gap identification for Case 3\label{fig:case3}}{%
  \includegraphics[scale=0.35]{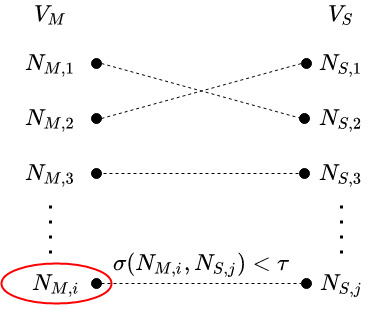}}
  \subcaptionbox{Gap identification for Case 4\label{fig:case4}}{%
  \includegraphics[scale=0.3]{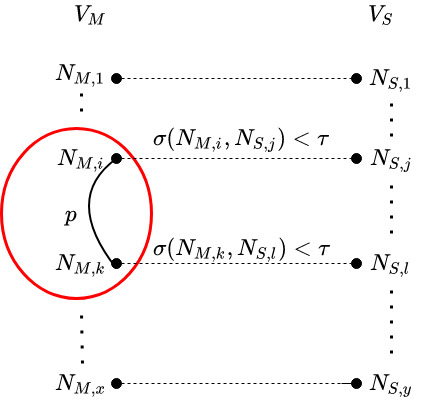}}
  \caption{Four different cases for gap identification from the derived alignment $\mathcal{X}$}
  \label{gap-cases}
\end{figure*}

While a subset of aligned node-pairs that survived the pruning step contains node-pairs from $\langle V_M, V_S\rangle$, there might be other nodes in $V_M$ or $V_S$ or both that do not participate in any alignment. These un-aligned nodes are the possible candidates of being gaps. 
Let the filtered alignment obtained between $G_M$ and $G_S$ using any of the filters (\textit{threshold}, \textit{exact} or \textit{best}) be denoted as $\mathcal{X}$ where $\mathcal{X}$ is either $\mathcal{T}$, $\mathcal{E}$ or $\mathcal{B}$. 
Four different cases are considered for inferring gaps out of the derived alignment $\mathcal{X}$. The cases have been visually illustrated in Figure \ref{gap-cases}. The cases have been formally presented using First-Order-Logic (FOL) language. Along with the \emph{Connect} predicate, the following FOL predicates are used to represent the cases.
\begin{itemize}
    \item $Edge(x,~p,~y)$ represents that fact the there is an edge from $x$ to $y$ with relation $p$.
    \item $GapNode(x)$ represents that a node $x$ is a gap.
    \item $GapEdge(x,~p,~y)$ represents that fact the subject ($x$), relation ($p$) and object ($y$) combine to form a gap.
\end{itemize}

\noindent \textbf{Case 1:} \textit{If a node $N_{M,i}$ in $G_M$ does not find any alignment in $G_S$ and does not have any neighbor, then $N_{M,i}$ is a gap.} This case is illustrated through Figure~\ref{fig:case1}. 
\begin{equation}
\begin{split}
    \forall_{i}~[\forall_j \neg Connect(N_{M,i},~N_{S,j})]~\wedge~
    [\forall_{k,p}\neg Edge(N_{M,i},~p,~N_{M,k})] \\
    \implies GapNode(N_{M,i})
\end{split}    
\end{equation}
\textbf{Case 2:} \textit{If nodes $N_{M,i}$ and $N_{M,k}$ in $G_M$ do not find any alignment in $G_S$, where $N_{M,i}, N_{M,k} \in V_M$ and $\langle N_{M,i}, p, N_{M_k}\rangle \in E_M$, then nodes $N_{M,i}$ and $N_{M,k}$ connected by edge $p$ is a gap.} Case 2 is illustrated in Figure \ref{fig:case2}.  
\begin{equation}
\begin{split}
    \forall_{i,k}~[\forall_{j,l} \neg Connect(N_{M,i},~N_{S,j})~\wedge~
    \neg Connect(N_{M,k},~N_{S,l})]~\wedge~\\
    [\exists_{p}~Edge(N_{M,i},~p,~N_{M,k})]  \implies GapEdge(N_{M,i},~p,~N_{M,k})
\end{split}    
\end{equation}

Case 3 and Case 4 are exclusively applicable for one-to-one alignment obtained using \textit{best filter} denoted by $\mathcal{B}$. \\
\noindent \textbf{Case 3:} \textit{If a node $N_{M,i}$ in $G_M$ finds an alignment in $G_S$ with $\sigma < \tau$ and does not have any neighbour, then $N_{M,i}$ is a gap.} Case 3 is illustrated in Figure \ref{fig:case3}.

\begin{equation}
\begin{split}
    \forall_{i}~[\exists_{j}~ Connect(N_{M,i},~N_{S,j})~\wedge~\sigma(N_{M,i},~N_{S,j})<\tau ~\wedge \\
    [\forall_{p,k}\neg Edge(N_{M,i},~p,~N_{M,k})] \implies GapNode(N_{M,i})
\end{split}    
\end{equation}

\noindent \textbf{Case 4:} \textit{If nodes $N_{M,i}$ and $N_{M,k}$ in $G_M$ find alignment in $G_S$ with $\sigma < \tau$, where $N_{M,i}, N_{M,k} \in V_M$ and $\langle N_{M,i},~p,~N_{M,k}\rangle \in E_M$, then nodes $N_{M,i}$ and $N_{M,k}$ connected by edge $p$ is a gap.} Case 4 is illustrated in Figure \ref{fig:case4}.   

\begin{equation}
\begin{split}
    \forall_{i,k}~[\exists_{j}~ Connect(N_{M,i},~N_{S,j})\wedge\sigma(N_{M,i},~N_{S,j})<\tau] \wedge\\
    [\exists_{l}~Connect(N_{M,k},~N_{S,l})\wedge\sigma(N_{M,k},~N_{S,l})<\tau] \wedge\\
    [\exists_{p}~Edge(N_{M,i},~p,~N_{M,k})] \implies GapEdge(N_{M,i},~p,~N_{M,k})
\end{split}    
\end{equation}

\begin{figure*}[ht]
\centering
\includegraphics[width=1.0\textwidth]{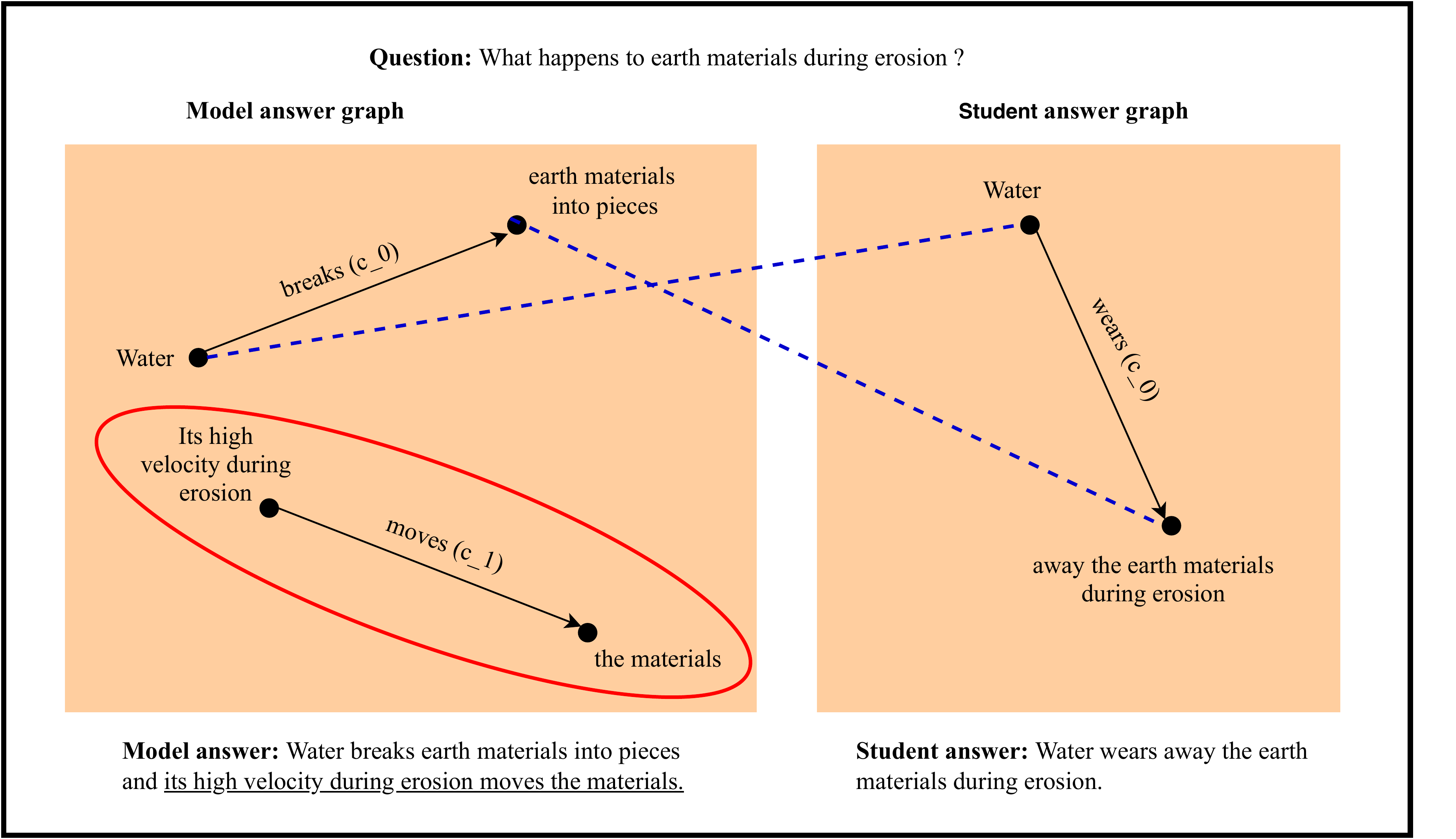}
\caption{Extraction of gap in student answer on alignment of answer graphs by \textit{FA-T}. Case 2 of gap identification strategy has been used here. }
\label{fig:four}
\end{figure*}

An example of gap extraction using Case 2 over FA-T model has been presented in Figure \ref{fig:four}. The nodes from model answer graph namely ``the materials'' and ``during erosion'' are not mapped to any node from student answer graph in the final alignment. As a result, the phrase ``the materials are moved during erosion'' is extracted as gap in student answer.  
\section{Experimental Design}
\subsection{Test Bed}
As an unsupervised approach towards gap extraction in the student answers has been adopted in this work, the development of gap annotated training data of adequate size can be skipped. However, a moderately sized gap annotated dataset is still required for validating the proposed unsupervised model. As the problem of gap extraction in the student answers have not been explored much, only a handful of datasets have been developed. The dataset developed by Dzikovska et al.\citep{dzikovska2012towards} annotates the answers with discrete categories, e.g., ``correct'', ``contradictory'', ``irrelevant'', etc. The annotations in this dataset are too coarse to provide meaningful feedback to the students. Godea et al.\citep{godea2016} addressed the issue by considering an annotation strategy that fragments the answers into a set of Minimal Meaningful Propositions (MMPs). The MMPs are classified into ``primary'', ``secondary'', ``extraneous'', and ``redundant'' categories. Bulgarov \& Nielsen \citep{bulgarov2018proposition} annotated the MMPs in the reference answer into ``understood'' or ``not understood'' categories. The MMPs are defined in sentence level as mentioned earlier. However, the gaps might surface in word or phrase levels as well. Nielsen et al.,\citep{Nielsen08} developed a very fine-grained annotation of entailment relationships in student answers. However, Bulgarov \& Nielsen \citep{bulgarov2018proposition} argued that ``facets are often too fine grained to be meaningful on their own''.  
The observations over the existing datasets motivated us to develop a new dataset by collecting question answer pairs from the three existing datasets, namely, University of North Texas (UNT)~\citep{mohler2009text}, SciEntsBank and Beetle \citep{dzikovska2012towards}, that are used as benchmark dataset in automated short answer grading domain. A subset of $\langle$question, model answer, student answer$\rangle$ triples from each dataset has been selected. 
The gaps in the student answers were marked by Subject matter Experts (SMEs) belonging to domains that are aligned to answers in the above mentioned three existing datasets such as Physics, General Science, Electronics, Computer Science. These SMEs are pursuing either PhD or have completed post-graduation in the above-mentioned domains. 

The SMEs who were involved in gap-annotation for the student answers especially from the University of North Texas dataset are PhD students from Centre for Educational Technology, Indian Institute of Technology Kharagpur (IIT Kharagpur) having good knowledge of Data Structures in Computer Science.

The SMEs involved in annotation of gaps in student answers for the SciEntsBank and Beetle datasets are PhD Research Scholars and postgraduate students from Centre for Educational Technology, IIT Kharagpur, having educational background and expertise in the field of Biological Sciences, General Science, Physics, Electronics Engineering. Such students are specifically chosen for gap-annotation process, as the SciEntsBank and Beetle datasets consisted of data pertaining to General Sciences, Physics and Electronics domain.

Table \ref{tab:ann} presents descriptive statistics of the annotated datasets of 926 student answers in response to 82 questions. 
$N$ represents total number of student answers in a particular dataset. $D$ indicates average gap density i.e., average number of gaps present in a student answer.  

\begin{table*}[ht]
\centering
\caption{Descriptive statistics of annotated datasets }
{\begin{tabular}{@{\extracolsep{\fill}}cccccc} \hline
Dataset & N & $Length_\mu$ & $Length_\sigma$ & $Length_{max}$ & D \\ \hline 
UNT & 161 & 25 & 14 & 94 & 0.68 \\ \hline
SciEntsBank & 418 & 13 & 7 & 54 & 0.902 \\ \hline
Beetle & 347 & 11 & 5 & 40 & 0.617 \\ \hline
\end{tabular}}
\label{tab:ann}
\end{table*}
In order to develop a gap-annotated dataset\footnote{Gap annotated dataset available at: \url{https://github.com/sahuarchana7/gaps-answers-dataset}} 
mentioned earlier, the manual annotation of gaps for each $\langle$Model answer, Student answer$\rangle$ pair has been performed in the following manner:

\begin{itemize}
    \item 
    Around 161-347 student answers are extracted from the set of all the student answers to the above-mentioned datasets as shown in Table \ref{tab:ann}. Table \ref{tab:ann} also provides statistics of the length of answers in each dataset. $Length_\mu$, $Length_\sigma$ and $Length_{max}$ indicate the mean / average number of words in an answer (mean / average length of the answer), the variability in the number of words in an answer and maximum number of words in an answer belonging to each of the above-annotated datasets, respectively.  It is to be noted that $Length_\mu$, $Length_\sigma$ and $Length_{max}$ values have been rounded off to the nearest integer for better understanding of the length of the answers and their variability in the datasets. The mean length of an answer varies from 11-25 words depending on the dataset, with the maximum answer length going up to 94 words and expected variability in the length of answer being 14 words from the mean for UNT dataset, for instance. Statistical values for other datasets are similarly shown in Table \ref{tab:ann}. 
    \item Each question may have a single model answer or multiple model answers. In case of multiple model answers, SMEs are assigned the task of aggregating multiple model answers into one relevant model answer. 
    \item SMEs compared the student answers with the corresponding model answers for a question and based on their knowledge and intuition, accordingly annotated the gaps in student answers through an annotation interface as shown in Figure \ref{interface}. 
    
    \begin{figure*}
    \centering
    \includegraphics[scale=0.4]{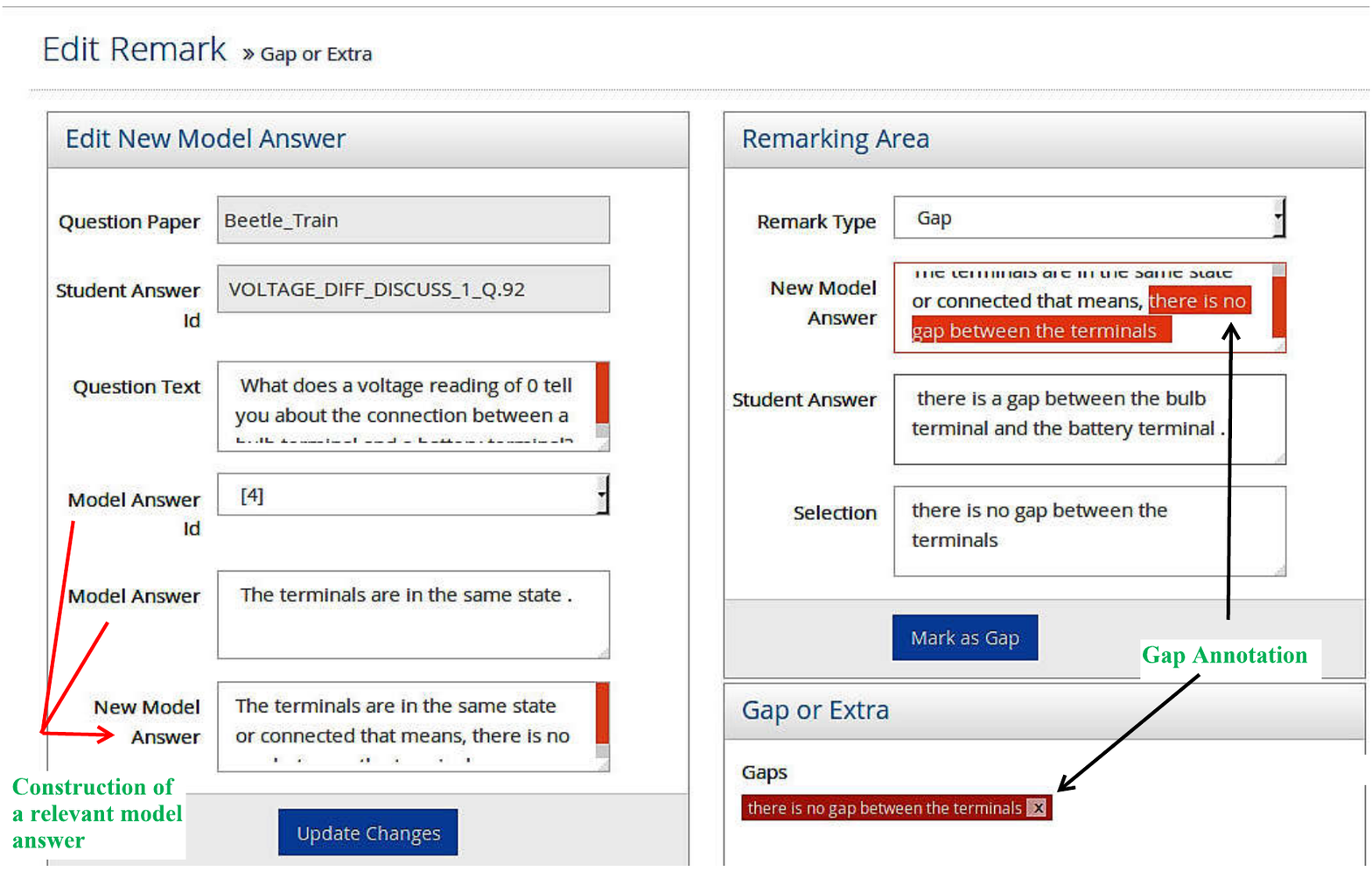}
    \caption{Different components of the gap annotation interface} 
    \label{interface}
    \end{figure*}

    The annotation interface consists of various fields namely:
    \begin{itemize}
    \item \textbf{Add/Edit new model answer:} This field consists of various sub-fields such as Question Paper, Student Answer Id, Question Text, Model Answer Id, Model Answer, New Model Answer. \\
    \emph{Question Paper} is selected by SMEs that contains the questions whose corresponding student answers would be marked for gaps. The specific ASAG domain dataset, mentioned earlier is indicated by the question paper. The student answer which is going to be annotated for gaps is referred to by \emph{Student Answer Id}. \emph{Question Text} indicates the question for the student answer to be gap-annotated. 
    \emph{Model Answer Id} and \emph{Model Answer} refer to the model answer(s) for the respective question, which may either be a single/multiple Model Answer Id/Model Answer according to question selected. A relevant model answer shown by \emph{New Model Answer} is constructed by visual observation of these model answer(s) by SMEs. The constructed relevant model answer is finalized with the help of \emph{Update Changes} button used by SMEs.

    \item \textbf{Area for remarks:} This field contains sub-fields such as Remark Type, New Model Answer, Student Answer, Selection. \emph{Remark Type} indicates the kind of remark (gap in the present task) to be assigned to the student answer. \emph{New Model Answer} and \emph{Student Answer} are reflected in this part of the interface. The SME has to select as gap a portion of the relevant model answer shown by \emph{New Model Answer}. The selected portion is reflected in \emph{Selection} area.  The SME has to click on \emph{Mark as Gap} button to finalize the gap according to him. 
    
    \item \textbf{Gap area:} The finalized gap annotated by SME, as described previously, is shown in this area of interface, indicating storage of this gap for the specific \emph{Student Answer Id} in database. 
\end{itemize}

\end{itemize}

An example of gap-annotation in student answers in the gap-annotated dataset is shown below:

\begin{mdframed}
\begin{Example}~\\
\label{ex:gap}
\textbf{Q:} How can a person test minerals for hardness using only the minerals (no tools)? \\
\textbf{M:} A person can test the minerals for hardness by rubbing the minerals together and \underline{seeing which one scratches the other}.\\
\textbf{S:} A person could rub the minerals together.\\
\textbf{Gap in S annotated by SME:} [`seeing which one scratches the other'] 
\end{Example}
\end{mdframed}



In traditional human annotation tasks, the statistics are computed and reported involving Inter-annotator agreement (IAA) using the chance corrected agreement measures like kappa coefficient. However, for the
developed gap annotation dataset, the IAA statistics based on kappa coefficient have not been reported owing
to certain issues that involve the nature of the present annotation task that treats the gap annotation as a sequence marking task where the sequence might be subject, object or an entire triple, in contrast to the existing datasets that treat the gap annotation in a classification setting.
Hence, in the present annotation task, there can often be multiple ways to annotate gap for a student answer. One
SME/annotator may annotate the entire phrase containing the keyword with supporting prepositions and helping verbs around the keyword (the entire triple with subject, object nodes and predicate showing some relation between the nodes instead of just the subject or object node), in the model answer as a
gap. On the other hand, another SME/annotator may feel that the single keyword without the supporting prepositions and helping verbs (subject or object node only), in the model answer is enough
to be annotated as a gap. Thus, the gaps annotated by different SMEs/annotators are of arbitrary granularity which may be
overlapping but of different length, as shown in Example \ref{ex:gap:comment}. Gap annotated by
SME1, i.e., [`is to move bones'] is longer than gap annotated by SME2, i.e., [`move bones']. 
\begin{mdframed}
\begin{Example}~\\
\label{ex:gap:comment}
\textbf{Q:} What is the main job of muscles in the body ? \\
\textbf{M:} The main job of muscles is to move bones.\\
\textbf{S:} The main job of muscles is to work together with the bones. \\
\textbf{Gap in S annotated by SME1:} [`is to move bones'] \\
\textbf{Gap in S annotated by SME2:} [`move bones']
\end{Example}
\end{mdframed}
Thus, there is no guarantee that every SME/ annotator will annotate
gaps in exactly the same terms. This poses a problem for IAA statistics such as kappa coefficient that rely on
an exact match to measure agreement between SMEs.\\
Consequently, we cannot use the inter-annotator agreement (IAA) measures like Kappa statistics to quantify the extent of agreement. Instead, we adopted a vetting-based strategy to verify the annotated gaps. For a given model answer and student answer pair, the gaps annotated by the first level annotator is verified by another second level annotator.

We observe that the second level annotators agreed by 78.3\%, 71\%, and 83.7\% with the first level annotators respectively for the UNT, Beetle and SciEntsBank dataset.   

It is to be noted that even though only the first level annotator marked the gaps in student answers, the second level annotator was also involved in the gap annotation process, as mentioned earlier, for verifying the gaps annotated by first level annotator.
In cases of disagreement between first level annotator and second level annotator, the annotations by first level annotator was taken into consideration, since the first annotator is an expert in the respective domain of the respective dataset. This process was followed to speed up the process of building a dataset with true gaps for the student answers. 
 
\subsection{Evaluation Metrics}
In \citep{arch2016identifygaps}, \textit{precision} has only been taken as the evaluation metric for measuring performance of formative feedback system. By drawing inspiration from the existing works \citep{bulgarov2018proposition}, we have used different variations of the Macro-average $F1$ measure to evaluate the proposed model. 

We define true gaps to be the gaps annotated by the annotator. Let the true gaps for a pair of answers $\langle M, S \rangle $ be denoted as \textbf{$True$} and $FA$ system prediction labels (gaps) be denoted as \textbf{$Sys$}, where:

$True = \{gp_1, gp_2, ...\}$, where $gp_1$, $gp_2$, ... indicate missing keywords/phrases of $M$ in $S$, known as gaps in $S$ in gap annotated data (by annotators). 

$Sys = \{sp_1, sp_2, ...\}$, where $sp_1$, $sp_2$, ... indicate missing keywords/phrases of $M$ in $S$, known as gaps in $S$ as predicted by the $FA$ system.

Let $\overline{Sys}$ indicates the set of keywords/phrases from $M$ which are not identified as gaps in $S$. For the $i^{th}$ student answer, the matching of $\overline{Sys_i}(j)$ (representing $j^{th}$ entry in  $\overline{Sys_i}$) with an element of $True_i$ has been done by maximal matching. F1 score of ROUGE-2 measure has been used to compute the extent of match between the set of the system predicted gaps ($Sys$) and that of the true gaps ($True$). For a gap in $Sys$, if the F1 score of ROUGE-2 with any of the gaps in $True$ exceeds a particular threshold ($\delta$), the gap is classified as true positive (TP)\footnote{It is to be noted that, for an $\langle M_i, S_i\rangle$ pair, once a system annotated gap $Sys_i(j)$ is matched with a true gap $True_i(l)$, that true gap is not considered for matching other system predicated gaps in $Sys_i$.} otherwise it is classified as false positive (FP). On the other hand, for a gap in $True$, if its F1 score of ROUGE-2 with all gaps in $Sys$ is below the threshold, the gap is grouped into the set of false negatives (FN). All the experiments are carried out with $\delta$ =0.5. 

\begin{equation}
    TP_{i} = \sum\limits_{j=1}^{\abs{Sys_{i}}} t
\hspace{0.1cm} ; \hspace{0.1cm} 
    t = 
    \begin{dcases}
    1 & \mbox{if } Sys_{i}(j) \in True_{i} \\
    0 & \mbox{otherwise}
    \end{dcases}
\end{equation}
\begin{equation}
    FP_{i} = \sum\limits_{j=1}^{\abs{Sys_{i}}} t
\hspace{0.1cm} ; \hspace{0.1cm} 
    t = 
    \begin{dcases}
    1 & \mbox{if } Sys_{i}(j) \notin True_{i} \\
    0 & \mbox{otherwise}
    \end{dcases}
\end{equation}

\begin{equation}
    FN_{i} = \sum\limits_{j=1}^{\abs{\overline{Sys_{i}}}} t
\hspace{0.1cm} ; \hspace{0.1cm} 
    t = 
    \begin{dcases}
    1 & \mbox{if } \overline{Sys_{i}}(j) \in True_{i} \\
    0 & \mbox{otherwise}
    \end{dcases}
\end{equation}

\noindent Let $TP_i$, $FP_i$ and $FN_i$ denote the true positive, false positive and false negative counts for a pair of answers $\langle M_{i}, S_{i} \rangle $. The $Precision_i$ and $Recall_i$ are computed using the $TP_i$, $FP_i$ and $FN_i$ counts in a standard manner. To report the performance of the proposed system, three measures have been used: 1) Question-based Macro-average Precision ($Macro\text{-}P_Q$), 2) Question-based Macro-average Recall ($Macro\text{-}R_Q$) and 3) Question-based Macro-average F1 ($Macro\text{-}F1_Q$).

Let $Precision_{q}$ and $Recall_{q}$ denote average precision and average recall computed over the set of student answers $S_q$ for a question $q \in Q$, where $Q$ is the set of all questions in dataset. 
\begin{equation}
Precision_{q} = \frac{1}{\abs{S_q}} \sum_{i=1}^{\abs{S_q}} Precision_{i}
\end{equation}

\begin{equation}
        Macro\text{-}P_{Q} = \frac{1}{\abs{Q}}\sum\limits_{q=1}^{\abs{Q}}Precision_{q} 
\end{equation}
The question-based recall ($Recall_q$) and the $Macro\text{-}R_Q$ are defined as follows:
\begin{equation}
   Recall_{q} = \frac{1}{\abs{S_q}} \sum_{i=1}^{\abs{S_q}} Recall_{i}
\end{equation}

\begin{equation}
        Macro\text{-}R_{Q} = \frac{1}{\abs{Q}}\sum\limits_{q=1}^{\abs{Q}}Recall_{q} 
\end{equation}
For a pair of answers $(M_{i}, S_{i})$ in dataset, F1 score ($F1_{i}$) is determined by harmonic mean of $Precision_{i}$ and $Recall_{i}$.
\begin{equation}
    \textit{F1 score}_{i} = \frac{2 \times Precision_{i} \times Recall_{i}}{Precision_{i}+ Recall_{i}}  
\end{equation}
The F1 score for a question $q\in Q$ is represented as:
  
\begin{equation}
    F1_{q} = \frac{1}{\abs{S_q}}\sum\limits_{i=1}^{\abs{S_q}}\textit{F1 score}_{i}
\end{equation}

\begin{equation}
        Macro\text{-}F1_{Q} = \frac{1}{\abs{Q}}\sum\limits_{q=1}^{\abs{Q}}\textit{F1 score}_{q}
\end{equation}

\subsection{Experimental Set-up}
In this work, we have presented three variations of the directed graph alignment-based approach towards gap identification: FA-T, FA-E and FA-B. The following experiments are conducted to measure the performance of the proposed models and compare them with System-I as baseline. Two-tailed t-test has been performed to check the statistical significance of the comparisons. 
\subsubsection{Experiment 1 - Dataset-wise Performance Analysis}
Evaluation metrics \textit{Macro-P}$_Q$, \textit{Macro-R}$_Q$ and \textit{Macro-F1}$_Q$ are computed for all formative assessment models against individual datasets. 
\subsubsection{Experiment 2 - Ablation Study highlighting impact of directionality in gap detection} 
An ablation study is conducted where for each dataset, we categorized the student answers into two groups: 
one group is comprised of student answers in which the directionality information in their respective answer graphs is of utmost importance for better understanding (\emph{Group-dir}) and the other group is comprised of student answers where the directionality information is not vital (\emph{Group-nodir}).

An illustration of a student answer belonging to group \emph{Group-dir} is shown below: 

\begin{mdframed}
\emph{Refining and Coding are influenced by testing stage.}
\end{mdframed}

\begin{figure}[h!]
 \centering
\includegraphics[scale=0.6]{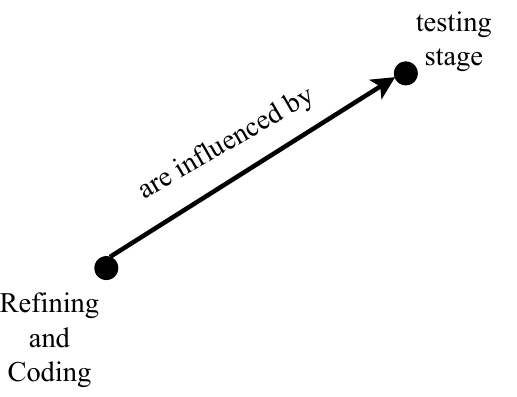}
\captionof{figure}{An illustrative example of a student answer belonging to group \emph{Group-dir} }
\label{group-dir}
\end{figure}

The triples for the above student answer are obtained as: \\
``Refining and Coding''   ``are influenced by'' ``testing stage'' \\
If the above set of triples are represented as an undirected student answer graph, the student answer could also be interpreted as ``Testing stage are influenced by refining and coding.'', which is actually not similar to the original student answer. Hence, the original student answer, as highlighted above in a box, is represented as a directed student answer graph as shown in Figure \ref{group-dir}. 

An illustration of a student answer belonging to group \emph{Group-nodir} is shown below: 
\begin{mdframed}
\emph{A stack is a finite ordered list with zero or more elements. } 
\end{mdframed}

The triples for the above student answer are obtained as: \\
``A stack''   ``is''    ``a finite ordered list with zero or more elements'' \\
The above triples could be used to build an undirected answer graph as shown below in Figure \ref{group-nodir}. 
\begin{figure}[h!]
 \centering
\includegraphics[scale=0.6]{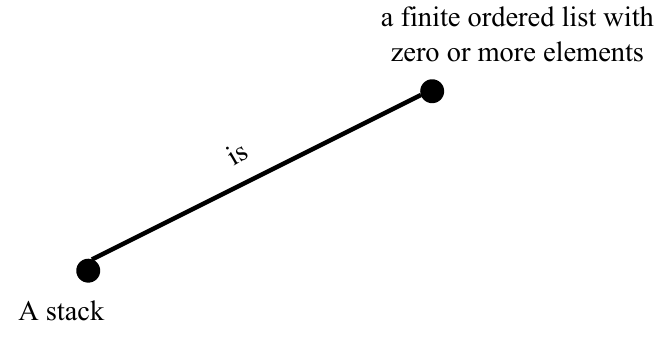}
\captionof{figure}{An illustrative example of a student answer belonging to group \emph{Group-nodir} }
\label{group-nodir}
\end{figure}
Since the edge between the nodes doesn't have any direction, as seen in the respective student answer graph in Figure \ref{group-nodir}, the respective student answer graph can be interpreted as ``A stack is a finite ordered list with zero or more elements'' or ``A finite ordered list with zero or more elements is a stack'' which have the same meaning. 

The performance of the proposed directed FA model and System-I are evaluated for each of the groups \emph{Group-dir} and \emph{Group-nodir}. 
\subsection{Experimental Results}
\subsubsection{Experiment 1} 
\label{datawise:comparison}
Table \ref{dw} depicts the comparison of different models over different datasets with the total number of student answers 
denoted by $N_T$ also being mentioned for each dataset. Model FA-T shows the best FA performance with $\textit{Macro-F1}_{Q}$ = 0.3089 (as compared to $\textit{Macro-F1}_{Q}$ = 0.3002 for System-I) for UNT dataset. In case of SciEntsBank (SciEnts) dataset, highest FA performance is reported by model FA-E with $\textit{Macro-F1}_{Q}$ = 0.3757 (in comparison to $\textit{Macro-F1}_{Q}$ = 0.3167 for System-I). However, for the Beetle dataset, maximum FA performance with $\textit{Macro-F1}_{Q}$ = 0.3893 is reported by System-I, where the proposed model FA-E shows reduced performance with $\textit{Macro-F1}_{Q}$ = 0.3537. 


\begin{table*}[ht]
\scriptsize
\centering
\caption{Dataset-wise performance analysis: $\textit{Macro-P}_{Q}$, $\textit{Macro-R}_{Q}$ and $\textit{Macro-F1}_{Q}$ are abbreviated as MAP, MAR and MAF1}
\label{dw}
\begin{tabular}{@{\extracolsep{\fill}}ccccccccccccc}
\hline
\multirow{2}{*}{} & \multicolumn{3}{c}{System-I} & \multicolumn{3}{c}{FA-B} & \multicolumn{3}{c}{FA-T} & \multicolumn{3}{c}{FA-E} \\ \cline{2-13} 
 & MAP & MAR & MAF1 & MAP & MAR & MAF1 & MAP & MAR & MAF1 & MAP & MAR & MAF1 \\ \hline
UNT ($N_{T}$=161)  & 0.3000 & 0.3053 & 0.3002 & 0.2742 & 0.3475 & 0.2992 & 0.3057 & 0.3152 & \textbf{0.3089} & 0.2852 & 0.3202 & 0.2967 \\ \hline
SciEnts ($N_{T}$=418) & 0.3160 & 0.3302 & 0.3167 & 0.3433 & 0.3728 & 0.3485 & 0.3209 & 0.3348 & 0.3236 & 0.3701 & 0.3946 & \textbf{0.3757} \\ \hline
Beetle ($N_{T}$=347) & 0.3830 & 0.4019 & \textbf{0.3893} & 0.3350 & 0.3780 & 0.3482 & 0.3352 & 0.3445 & 0.3383 & 0.3445 & 0.3777 & 0.3537 \\ \hline
\end{tabular}
\end{table*}
\begin{figure}[h!]
\centering
\includegraphics[scale = 0.6]{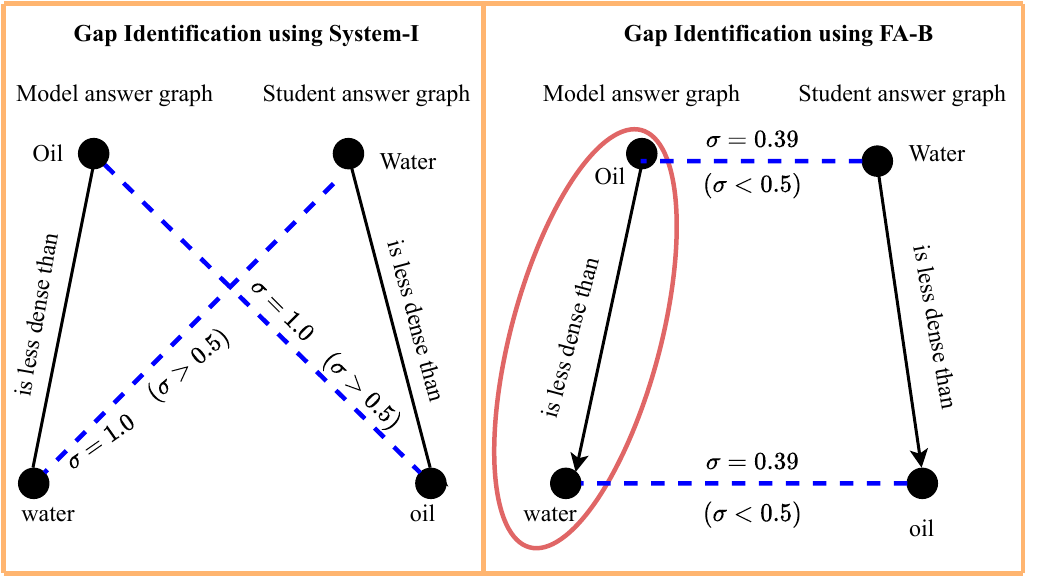}
\caption{Impact of directionality in gap identification}
\label{fig:comparison}
\end{figure}

\subsubsection{Experiment 2}
Table \ref{as} presents the number of answers in each of the groups \emph{Group-nodir} denoted as $N_{nodir}$ and \emph{Group-dir} denoted as $N_{dir}$ belonging to each dataset as well as the System-I performance and maximum performance amongst all proposed directed FA models FA-T, FA-B, FA-E in each case. 


\subsection{Analysis of Experimental Results}
From almost all the comparisons, it is observed that the \emph{exact} and the \emph{best} filter-based models (e.g., FA-E and FA-B) perform better than the \emph{threshold} filter-based models e.g., FA-T and System-I. It is to be recalled that both FA-E and FA-B consider one-to-one alignment between the nodes in a model answer graph and a student answer graph whereas the FA-T and System-I model consider a many-to-many alignment. Due to the relaxation in alignment allowance, the nodes that are indicative of gaps might find alignment in FA-T or System-I. Consequently, these models fail to mark these nodes as gaps. On the other hand, the FA-E and FA-B models, due to their stringent one-to-one alignment strategy, may leave the gap nodes unaligned which in turn facilitates discovery of true gaps. 

\begin{table*}[ht]
\centering
\caption{Ablation Study: Dataset-wise performance analysis: $\textit{Macro-F1}_{nodir}$ and $\textit{Macro-F1}_{dir}$}
\label{as}
\begin{tabular}{@{\extracolsep{\fill}}ccccc}
\hline
\multirow{2}{*}{} & \multicolumn{2}{c}{System-I} & \multicolumn{2}{c}{Highest Proposed Method Performance} 
\\ \cline{2-5} 
 & $\textit{Macro-F1}_{nodir}$ & $\textit{Macro-F1}_{dir}$ & $\textit{Macro-F1}_{nodir}$ & $\textit{Macro-F1}_{dir}$  \\ \hline
UNT \\ ($N_{nodir} = 8, N_{dir} = 153$) & 0.1111 & 0.265 & 0.3278 & 0.2955 \\ \hline
SciEnts \\ ($N_{nodir} = 8, N_{dir} = 410$) & 0.33 & 0.314 & 0.48  & 0.3718 \\ \hline
Beetle \\ ($N_{nodir} = 133, N_{dir} = 214$) & 0.0383 & 0.060 & 0.1273 & 0.2195 \\ \hline
Average performance \\ over all datasets & 0.160 & 0.2132 & 0.312 & 0.2956 \\ \hline
\end{tabular}
\end{table*}


In the example presented in Figure~\ref{fig:comparison}, the FA-B model ensures appropriate (subject-subject) and (object-object) alignment guided by the directionality of the edge ``is less dense than''. Consequently, the true gap is detected by the proposed directed graph model. 
All the datasets comprise of comparatively more number of student answers in \emph{Group-dir} as compared to that in \emph{Group-nodir} which leads to maximum use of Proposed directed FA models for enhanced performance especially for answers that involve vital information which can only be interpreted with directionality in answer graphs. \\
It is observed from the results of ablation study as shown in Table \ref{as}, that for the student answers belonging to both the groups \emph{Group-dir} and \emph{Group-nodir} in case of all datasets, the highest performance amongst all proposed directed FA models FA-T, FA-B, FA-E exceeds the performance of System-I. The same is observed from the average model performance over all datasets with  $\textit{Macro-F1}_{Q}$ = 0.2956 that outperforms System-I (that shows $\textit{Macro-F1}_{Q}$ = 0.2132), for \emph{Group-dir}. At the same time, even for \emph{Group-nodir}, the proposed directed FA model is a clear winner showing $\textit{Macro-F1}_{Q}$ = 0.312 (as compared to $\textit{Macro-F1}_{Q}$ = 0.160 
for System-I).  

It is also observed that System-I has performed better than the proposed model on the Beetle dataset. A closer look at the structure of the answers in this dataset revealed that the number of answer-pairs containing only auxiliary verbs such as ``is'', ``have'', etc. and no action verbs, is comparatively higher than that in the SciEntsBank and UNT datasets. Thus, due to the less variety of verbs (predicates) in the answer-pairs in Beetle, the corresponding pair of canonical answer graphs would contain very few node-pairs that are a part of predicate induced alignment in the proposed directed FA model. As a result, very few node-pairs in the pair of canonical answer graphs would have neighbors to contribute to their initial similarity values. This leads to the fixpoint scores being highly dependent on the initial similarity values in most cases. In contrast, System-I, essentially considers support from all the neighboring node-pairs towards computation of similarity score of the concerned node-pair. This indicates substantial contribution of the neighborhood similarity scores towards the computation of similarity score of a node-pair, in System-I, as compared to that in the proposed directed FA model. The resulting alignment of answer graphs and eventually, performance with regards to extraction of gaps, is hence expected to be better in System-I in comparison to that in the proposed directed FA model. This is clearly evident from Table \ref{dw} that System-I has shown maximum FA performance with $\textit{Macro-F1}_{Q}$ = 0.3893 as compared to the best directed graph-based model FA-E with $\textit{Macro-F1}_{Q}$ = 0.3537. It is to be noted that the code for all the implementation discussed in this paper is publicly available at the location \url{https://github.com/sahuarchana7/Code-and-Dataset-for-Directed-Graph-Alignment-Approach-for-Student-Answers}.
In the next section, different issues that have impact on the performance of the proposed FA models are discussed.
\section{Discussion}
\subsection{Shortcoming due to Information Extraction (IE) Tool}
The proposed FA model uses an Information Extraction (IE) tool for extraction of triples from a pair of answers. It is observed that, in a number of cases, the triples generated using the IE tool seem inaccurate and redundant. This sometimes leads to the answer graphs that are unnecessarily large and noisy in nature. This inconsistency cascades to the similarity flooding and subsequently to the gap extraction steps. As a result, the gaps detected in student answer become improper or unavoidable.
An illustration concerning this inconsistency based on Example  \ref{ex:ie1} is presented in Table \ref{ie:shortcoming2}. \\
It is observed from Table \ref{ie:shortcoming2}, that the number of redundant gaps detected by directed graph-alignment based FA model while using triples generated using IE tool, are more in number in comparison to a case where manually-generated triples for constructing the answer graphs. The concerned FA model is thus expected to show improved performance using triples for a pair of answers, that appear more meaningful than triples obtained using IE tool. To investigate the general trend concerning this observation, triples related to the sentences from a set of randomly chosen 130 answers (from 17 questions in the SciEntsBank dataset) were manually extracted.  It is observed that there is a considerable improvement in the performance of the FA-E model with manually extracted triples ($\textit{Macro-F1}_{Q}$ = 0.3860) over the FA-E model with the automatically extracted triples  ($\textit{Macro-F1}_{Q}$ = 0.3143). The performance is expected to have similar trends in the other datasets. \\
\begin{mdframed}
\begin{Example} ~\\
\label{ex:ie1}
\textbf{Q:} What is a variable? \\
\textbf{M:} A variable is a location in memory that can store a value. \\
\textbf{S:} A variable is the memory address for a specific type of stored data, or from a mathematical perspective, a symbol representing a fixed definition with changing values. \\
\textbf{True gaps:} [ ] 
\end{Example}
\end{mdframed}
\subsection{Large Difference in Answer Pair Size}
There are cases where the student and model answers differ with respect to size to a very large extent. Consequently, the difference in the size of the corresponding answer graphs is large as well. In this setting, the Similarity Flooding Algorithm fails to find sufficient context for alignment. An exhibit of the present case in depicted in Example~\ref{ex:ex-short-ans}. 
\begin{mdframed}
\begin {Example} ~\\
\label{ex:ex-short-ans}
\textbf{Q:} What is the advantage of linked lists over arrays? \\
\textbf{M:} The linked lists can be of variable length. \\
\textbf{S:} One can insert elements into a linked list at any point . One does not need to resize linked list unlike one needs to resize an array . \\
\textbf{True gaps:} [ ] \\
\textbf{System detected gaps:} [of variable length] 
\end{Example}
\end{mdframed}
\begin{table*}[h!]
\scriptsize
\centering
\caption{Comparison of gaps extracted with triples generated using IE tool and extracted manually}
\begin{tabular}{@{\extracolsep{\fill}}ccc}
\hline
 & Triples extracted using IE tool & Triples extracted manually \\ \hline
M & \begin{tabular}[c]{@{}c@{}}\textless{}A variable, \quad is, \quad a location in memory\textgreater\\ \textless{}a location in memory, \quad can store, \quad a value\textgreater\\ \textless{}A variable, \quad is, \quad a location\textgreater{}\end{tabular} & \begin{tabular}[c]{@{}c@{}}\textless{}A variable, \quad is, \quad a location in memory\textgreater\\ \textless{}A location in memory, \quad can store, \quad a value\textgreater{}\end{tabular} \\ \hline
S & \begin{tabular}[c]{@{}c@{}}\textless{}A variable, \quad is, \quad the memory address\\  for a specific type of stored data\\  or from a mathematical perspective\textgreater  \\ \textless{}a symbol, \quad be representing, \quad a fixed definition\\  with changing values\textgreater\\ \textless{}A variable, \quad is, \quad the memory address\textgreater\\ \textless{}a mathematical perspective, \quad is, \quad the symbol\\  representing a fixed definition with changing\\ values\textgreater{}\end{tabular} & \begin{tabular}[c]{@{}c@{}}\textless{}A variable, \quad is, \quad the memory address\\  for a specific type of stored data\textgreater\\ \textless{}A variable, \quad is, \quad from a mathematical \\ perspective, a symbol\\ representing a fixed definition with \\ changing values\textgreater\\ \textless{}A symbol, \quad representing, \quad a fixed definition \\ with changing values\textgreater{}\end{tabular} \\ \hline
\begin{tabular}[c]{@{}c@{}}Gaps extracted in S \\ by directed \\ graph-alignment \\ based FA model \end{tabular} & {[}a location, a value{]} & {[}a value{]} \\ \hline
True gaps & - & - \\ \hline
\end{tabular}
\label{ie:shortcoming2}
\end{table*}
\section{Conclusion}
In this work, a directed graph alignment approach towards identifying gaps in a student answer has been presented. 
The proposed system recognizes the portions of the model answer that do not directly align with the input student answer. It is to be noted that the proposed FA model does not attempt to recognize errors in the student answer that don't directly align with M.
The existing body of work in the FA problem domain has addressed the problem by considering the gaps or feedback in a relatively fixed granularity (word level or propositional) level. The present work discussed in the paper attempted to bring in a flexibility in defining gaps at varying granularity (subject, object or combination). In order to bring in the required level of abstraction, the present work contributes primarily by defining the problem of automatic gap annotation in a formal graph theoretic framework. The formal graph theoretic problem abstraction has led to the development of a novel unsupervised approach towards annotating gaps. We believe that this is one of the few attempts concerning an unsupervised approach for the automatic gap identification. \\
The present work has also contributed by developing a publicly available dataset where the gap annotation reflects the flexibility in marking the gaps. This dataset has been used to measure the performance of three model variations, namely, FA-T, FA-E and FA-B. The performance of the proposed models have been compared against a strong baseline that considers the undirected graph-alignment based modelling. 
A statistically significant improvement in the performance with respect to the baseline system has been observed. Overall, the performance of the proposed system is promising considering the flexibility in annotating gaps. The proposed system has a few limitations: 1) concerning the extraction of appropriate triples; and 2) the incorrect alignment of graphs for the longer student answers. These limitations provide future scopes of work in this problem domain.

\newpage
\renewcommand{\thesubsection}{\Alph{subsection}}
\appendix
\section*{Appendix}
\subsection{Notations}
\begin{flushleft}
\begin{longtable}{lp{5.3cm}}
$Q$ &       Question\\ \\
$M$ &       Model Answer \\ \\
$S$ &       Student Answer \\ \\
$M_{t}$&       Triple set for Model Answer, i.e. set of all the triples that are extracted from the model answer $M$ and is of the form $\langle s, p, o \rangle$ \\ \\
$s$&        subject\\ \\
$p$&        predicate\\ \\
$o$&        object\\ \\
$\mathcal{S}_M$&        the set of subjects belonging to the triples in $M_{t}$ that have been extracted from $M$ using an Information Extraction (IE) tool.  \\ \\
$\mathcal{O}_M$&        the set of objects belonging to the triples in $M_{t}$ that have been extracted from $M$ using an Information Extraction (IE) tool.   \\  \\
$\mathcal{P}_M$&        the set of predicates belonging to the triples in $M_{t}$ that have been extracted from $M$ using an Information Extraction (IE) tool. \\  \\
$S_{t}$&      Triple set for Student Answer, i.e. set of all the triples that are extracted from the student answer $S$ and is of the form $\langle s, p, o \rangle$ \\ \\
$\mathcal{S}_S$&        the set of subjects belonging to the triples in $S_{t}$ that have been extracted from $S$ using an Information Extraction (IE) tool.  \\   \\
$\mathcal{O}_S$&        the set of objects belonging to the triples in $S_{t}$ that have been extracted from $S$ using an Information Extraction (IE) tool.   \\  \\
$\mathcal{P}_S$&        the set of predicates belonging to the triples in $S_{t}$ that have been extracted from $S$ using an Information Extraction (IE) tool. \\  \\
$G_M$&        Denoting Model Answer Graph in the mathematical representation of FA problem as well as Canonical Model Answer Graph in the proposed FA model for the sake of simplicity \\ \\
$G_S$&        Student Answer Graph\\ \\
$V_M$&        Set of vertices in $G_M$ obtained by $\mathcal{S}_M \cup \mathcal{O}_M$ \\ \\
$V_S$&      Set of vertices in $G_S$ obtained by $\mathcal{S}_S \cup \mathcal{O}_S$ \\ \\
$E_M$&      Set of directed edges in $G_M$ that has one-to-one correspondence with $\mathcal{P}_M$\\ \\
$E_S$&      Set of directed edges in $G_S$ that has one-to-one correspondence with $\mathcal{P}_S$\\ \\
Information
Extraction tool ClausIE &      ClausIE-Clause-Based
Open Information Extraction: databases-and-information-systems/software/clausie/
https://www.mpi-inf.mpg.de/departments/ \\ \\
$\sigma (x, y)$&    Similarity between contents of nodes in a pair $(x, y)$, where $x\in V_M$ and $y \in V_S$\\ \\
$\sigma^{o}(x, y)$&       Initial alignment score for the node-pair $(x, y)$ \\ \\
$\sigma^{c}(x, y)$&       Final alignment score obtained after convergence of the directed graph-alignment algorithm for the node-pair $(x, y)$ \\ \\
$\sigma_{f}$&       Maximum of $\sigma^{c}(x, y)$ and $\sigma^{o}(x, y)$ for each node-pair $(x, y)$ \\ \\     
$\Phi_k$&      A one-to-one alignment between answer graphs $G_M$ and $G_S$ such that  each node of $G_S$ is aligned to at most one node
of $G_M$ .  \\ \\
$\Phi^{*}$&      Optimal one-to-one alignment obtained between answer graphs $G_M$ and $G_S$ using \textit{best filter} which is $\Phi_k$ having the maximum alignment score  \\ \\
$\mathcal{T}$&      Set of node-pairs obtained using the threshold filter for a pair of answer graphs $\langle G_M, G_S \rangle$ which represents the best matches obtained after pruning the node-pairs having similarity value below a particular threshold. \\ \\
$\mathcal{E}$&      Set of best matches for a pair of answer graphs $G_M$ and $G_S$ obtained using \textit{exact filter}  \\ \\
$\mathcal{X}$&      Filtered alignment obtained between $G_M$ and $G_S$ using any of the filters (\textit{best} or \textit{threshold} or \textit{exact}), i.e., either $\mathcal{B}$, $\mathcal{T}$ or $\mathcal{E}$ \\ \\
$G(\mathcal{Y})$&       The set of gaps detected in $S$ based on filtered alignment $\mathcal{Y}$ \\ \\
$\textit{FA-T}$&       Gap detection model using the threshold filter \\ \\
$\textit{FA-E}$&       Gap detection model using the exact filter \\ \\
$\textit{FA-B}$&       Gap detection model using the best filter \\ \\ 
$\tau$&     Threshold set apriori for node-pair similarity in the proposed FA model \\ \\
\end{longtable}
\end{flushleft}



%

\subsection{Answer-Graph Node Representation}
The predicate is a word / phrase that can be converted to a vector using CBOW model. 
Continuous Bag-of-word (CBOW ) model is a context predictive model implemented with Word2vec \citep{rong2014word2vec}. The corpus to train the word vectors has been created by collecting sentences belonging to a set of Wikipedia articles that are related to the text in the three grading/labeling corpus, namely, UNT, SciEntsBank and Beetle. These Wikipedia articles are extracted from the Wikipedia database in the following manner:
\begin{itemize}
\item For a pair of question and corresponding model answer in the grading/labeling corpus, all the content words are identified, and the Wikipedia pages related to these words are retrieved using the TAGME service\footnote{\url{https://tagme.d4science.org/tagme/i}} . Set of Wikipedia articles thus obtained are called the first level articles.
\item The Wikipedia articles that are hyperlinked from the definition blocks of the first level Wikipedia articles are collected further and are added to the corpus.
\item The above steps are performed for pairs of all questions and corresponding model answers in the grading/labeling corpus leading to extraction of a large number of Wikipedia articles. The total number of Wikipedia articles that were collected in two passes are 21,265.
The trained CBOW model is used to construct word vectors for each of the words in a predicate.  The mean of the word vectors corresponding to each of the words in the predicate indicates the word vector for the predicate. 
\end{itemize}

The number of OOV (Out-of-Vocabulary) words is expected to be very small, as we have tried to build the corpus by fetching all possible Wikipedia articles belonging to the content words in the grading/labeling corpus. Even then , if an OOV term appears we have ignored it in the present work. In future, FastText vector building process could be used instead of  CBOW model from Word2vec , so that OOV terms are handled in a better manner. 

\section*{Competing interests:} The author(s) declare none.

\bibliographystyle{unsrtnat}
\bibliography{references}  






\end{document}